\documentclass[11pt,a4paper]{article}
\usepackage[margin=25mm]{geometry}
\usepackage[T1]{fontenc}
\usepackage[utf8]{inputenc}
\usepackage{lmodern}
\usepackage{amsmath,amssymb,graphicx,booktabs,longtable,multirow,array,calc}
\usepackage{microtype}
\usepackage{caption}
\usepackage[section]{placeins}
\usepackage{xurl}
\usepackage[hidelinks,unicode]{hyperref}
\title{When Position Favors the Wrong Part: Evaluating Left-Right Discrimination in Semantic Correspondence}
\author{Pingan Wang\\[4pt]
\small College of Artificial Intelligence and Low-Altitude Technology,\\
\small South China Agricultural University, Guangzhou, 510642, China\\[3pt]
\small\texttt{pinganwang@scau.edu.cn}}
\date{}
\hypersetup{pdftitle={When Position Favors the Wrong Part: Evaluating Left-Right Discrimination in Semantic Correspondence},pdfauthor={Pingan Wang},pdfcreator={pdfLaTeX}}
\begin{document}
\maketitle
\begin{abstract}
Overall part-discrimination accuracy can conceal failures when image position favors the wrong bilateral part. We examine whether dense descriptors distinguish the correct part under this conflict and whether their discrimination scores help select matching coordinates. We evaluate position agreement and conflict within the same target images and annotated part pairs, using a bounding-box-side score to define the conditions. We compare the original descriptors with their even and odd components under horizontal reflection, examine feature upsampling with the backbone and adapters fixed, and test candidate-selection rules based on odd scores. On SPair-71k, odd discrimination accuracy falls from 86.66\% under agreement to 19.63\% under conflict for frozen DINOv2, compared with 95.27\% to 86.16\% for MARCO. With MARCO's input geometry fixed, odd accuracy in box-side conflict was 34.12\% on the native grid and 88.84\% after learned upsampling and pooling back to that grid. Odd-score rules improve on the original hard-peak matchers overall for DINOv2 and DINOv3 and on conflict queries for MARCO. With MARCO's soft decoder fixed, reranking the cosine peaks changes mean per-pair PCK by +0.47 pps. For supervised GeoAware-SC, using the odd score to select the localization window center yields only 15 net additional correct matches, with 330 corrections and 315 introduced errors. Evaluating discrimination under position conflict reveals failures hidden by aggregate accuracy. Reporting corrections and introduced errors separately also shows whether a discriminative score produces useful matching decisions.
\end{abstract}

\section{Introduction}
\label{sec:introduction}

Consider a model asked to match a left eye across two photographs. It must find an eye in the target image and determine which eye it is. If the object changes orientation, the correct eye can appear on opposite sides of the two images. A match based on image position can then select the other eye.

\begin{figure}[!htbp]
\centering
\includegraphics[width=\linewidth,height=0.77\textheight,keepaspectratio]{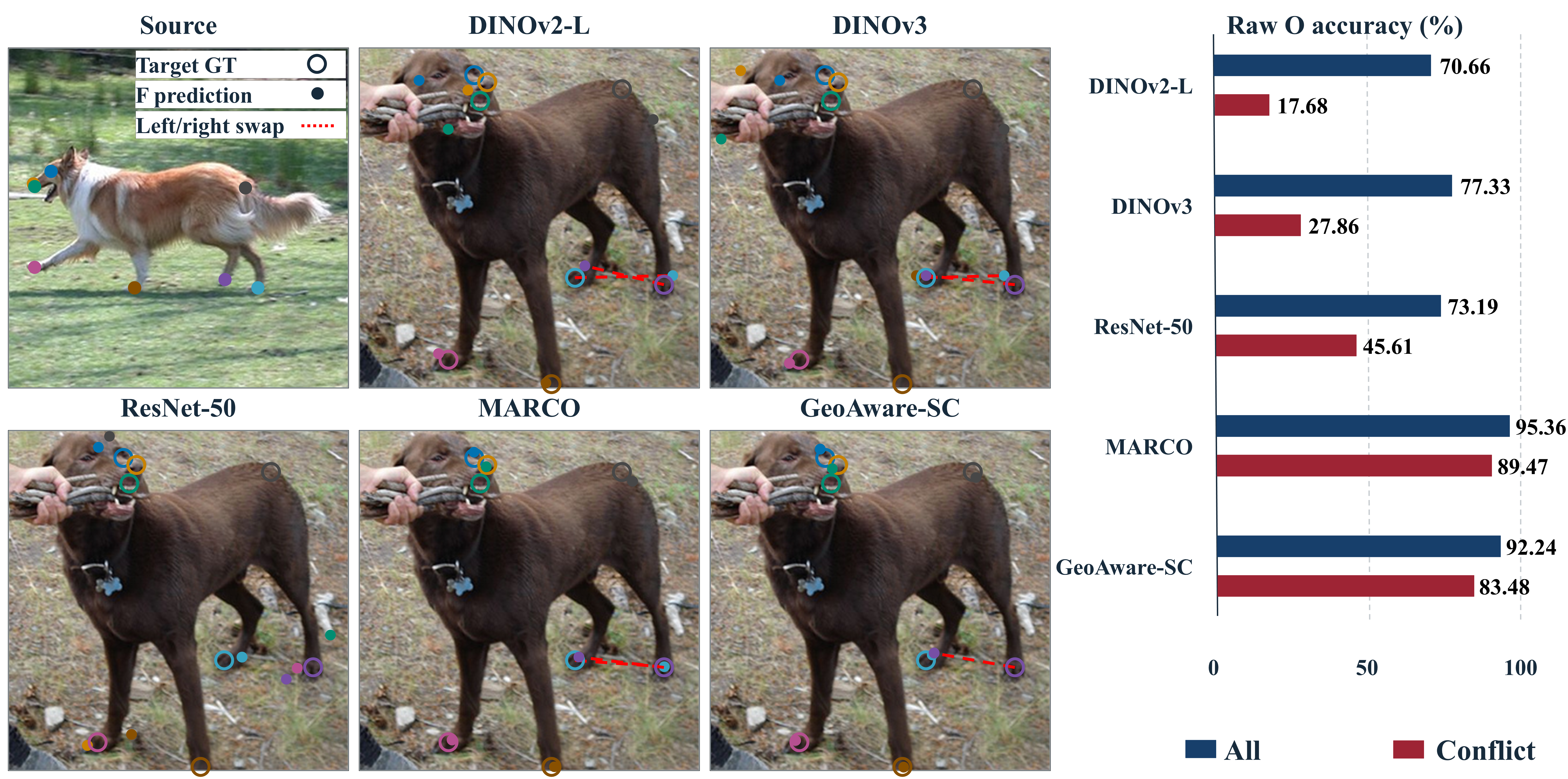}
\caption{Left/right matching errors across five models. The bars show raw odd discrimination on all eligible queries and on box-side conflicts.}
\label{fig:1}
\end{figure}

Self-supervised models provide transferable dense descriptors \cite{r1,r2}. Correspondence adaptation \cite{r3} and pose alignment \cite{r4} improve matching. In our SPair-71k evaluation \cite{r5}, left/right swaps account for 32.0\% of the original matching errors for frozen DINOv2 and 25.9\% for MARCO. Figure~\ref{fig:1} illustrates such errors. Overall discrimination accuracy can conceal differences between queries where the position cue favors the correct part and those where it favors its counterpart. On the eligible queries, the odd component of DINOv2 reaches 70.66\%, close to the 70.26\% of a score based on horizontal positions within annotated boxes (Table~\ref{tab:1}).

We ask three questions. First, can a descriptor distinguish the correct part from its counterpart when a position cue favors the counterpart? We compare the source descriptor's similarity to each of the two annotated target parts using the original features and the even and odd components of their aligned responses to the original and reflected images. We group queries by whether position favors the correct part, favors its counterpart or gives both the same score. Within each target image and part pair, we then compare queries for which position favors opposite choices.

Second, how does MARCO's learned upsampling head affect part discrimination under position conflict? Holding MARCO\textquotesingle s backbone and adapters fixed, we compare four readouts of the exact same adapted 60 \(\times\) 60 feature map: the native 60 \(\times\) 60 grid, bilinear interpolation to 240 \(\times\) 240, the learned 240 \(\times\) 240 upsampling head, and the learned 240 \(\times\) 240 output pooled back to 60 \(\times\) 60. We also test sensitivity to image placement and to positional-embedding or patch-kernel symmetrization.

Third, even when a descriptor distinguishes two annotated target parts more accurately, do those discrimination scores help a matcher choose the correct coordinate across the full target image? We test fixed candidate-selection rules across both hard-peak matchers and soft-localization decoders, compare the resulting percentage of correct keypoints (PCK) against position baselines and inference controls, and count corrected errors and newly introduced errors separately alongside sign-reversal and equal-count switching controls.

We make three contributions.

\begin{enumerate}
\def\labelenumi{(\arabic{enumi})}
\item
  We propose evaluating left-right part discrimination under conflicting position cues. Comparisons within the same target images and annotated part pairs show that aggregate accuracy conceals substantial differences between the tested models when position favors the annotated bilateral counterpart.
\item
  We compared four readouts of the same MARCO features with the backbone and adapters fixed. The learned upsampling head improved left/right discrimination under position conflict for both the full descriptor and its odd component. Most of this gain remained when we pooled the head's 240 \(\times\) 240 output back to the native 60 \(\times\) 60 grid. Bilinear interpolation of the native features alone had little effect.
\item
  We test the use of odd scores in candidate selection. By counting both corrected and introduced errors, we show where the scores improve matching and identify cases where better discrimination accompanies little net change in matching.
\end{enumerate}

Figure~\ref{fig:1} shows the same eight source queries and the target locations predicted by five original matchers. Open rings mark target ground truth, filled points mark predictions, and dashed red segments mark left-right swaps. The bars report raw odd accuracy overall and on the subset for which the box-side cue favors the counterpart. Reading the panels together motivates two separate tests: how well a descriptor distinguishes annotated parts, and whether it helps a matcher select the correct coordinate.

\section{Related Work}
\label{sec:related-work}

\subsection{Dense Features and Correspondence}
\label{sec:dense-features-and-correspondence}

Proposal Flow matches appearance and geometry \cite{r6}. Later methods learn dense scores \cite{r7}, combine diffusion and self-supervised features \cite{r8}, and use pseudo labels \cite{r9}, 3D priors \cite{r10} or correspondence adaptation \cite{r3}. Unified benchmarking compares extraction conventions \cite{r11}; PDC-Net estimates flow uncertainty \cite{r12}, and RoMa learns two-view matches \cite{r13}. Other approaches use keypoint prototypes \cite{r14}, consistent features across images \cite{r15}, 3D-guided cycle consistency \cite{r16}, or query expansion for object retrieval \cite{r17}. Wang et al. examine how feature aggregation degrades spatial structure \cite{r18}.

\subsection{Bilateral Ambiguity and Chirality}
\label{sec:bilateral-ambiguity-and-chirality}

Viewpoint reasoning connects orientation and part labels~\cite{r19}. GeoAware-SC addresses DINOv2's left/right difficulty through dense training and pose alignment~\cite{r4}. Spherical maps~\cite{r20} and foundation-model probes~\cite{r21} examine related geometry.

Visual chirality concerns changes in the distribution of natural images under reflection \cite{r22}. \(\chi\) aligns features from original and reflected images, aggregates them on 3D shapes, and learns chirality descriptors \cite{r23}. Wei\ss berg et al. learn symmetry-informative and symmetry-agnostic descriptors for 3D shape analysis \cite{r24}. Pix2LR learns pixel-level left/right predictions and evaluates side labels at annotated keypoints on SPair-71k and PF-PASCAL \cite{r25}. We ask whether a source descriptor favors the correct target part over its counterpart when position scores favor the counterpart. We compare these preferences within the same target images and part pairs, then count how often using the scores to choose a target coordinate corrects or introduces a matching error.

\subsection{Equivariance and Position Cues}
\label{sec:equivariance-and-position-cues}

Group-equivariant networks~\cite{r26}, empirical equivariance tests~\cite{r27}, patch-sampling studies~\cite{r28}, and reflection-aware training~\cite{r29} examine transformation responses. Yamamoto et al. study position/token interactions in a synthetic CLIP-style setting~\cite{r30}. Denoising Vision Transformers~\cite{r31} and registers~\cite{r32} address spatial artifacts; Positional Encoding Image Prior examines position in reconstruction~\cite{r33}. We compare position scores with features extracted from natural images using models that include Contrastive Language-Image Pretraining (CLIP)~\cite{r34} and ResNet-50~\cite{r35}. In INSID3, Cuttano et al. \cite{r36} project DINOv3 features away from an estimated positional subspace. On SPair-71k, overall PCK@0.10 for DINOv3-L rises from 50.6\% to 52.0\%. Our comparison asks which of two annotated parts the descriptor favors when position points to the wrong one.

\section{Materials and Methods}
\label{sec:materials-and-methods}

We first compare left/right discrimination within the same target image and annotated part pair, grouping source queries by which part the position cue favors. To see how the feature readout affects this comparison, we hold the MARCO backbone and adapters fixed and compare native, interpolated and learned features at matched grid sizes. We then use the odd scores to select among candidate matches and count how often a different choice corrects an error or introduces one.

\subsection{Measures of Discrimination and Matching}
\label{sec:measures-of-discrimination-and-matching}

The discrimination test asks whether a descriptor can distinguish the correct part from its counterpart when both target locations are given, including when position favors the counterpart. In the matching test, the matcher searches the target image, and PCK checks whether its predicted coordinate is within tolerance of the annotation. On SPair-71k, discrimination is evaluated on 28,978 eligible queries, and overall PCK uses all 88,328 visible queries. Let \(I_{s},I_{t}\) denote the source and target images, \(q\) the source query, and \(F(I)(p)\) the normalized descriptor sampled at location \(p\). For any descriptor \(G\), the similarity between the source query and a target location is

\begin{equation}
s_{G}(p \mid q) = \cos\left( G(I_{s})(q),G(I_{t})(p) \right).
\label{eq:1}
\end{equation}

Cosine scoring normalizes both operands and assigns zero when either has zero norm. We evaluate \(G = F,E,O,O_{res}\). For an annotated query, \(p^{+}\) is the true target and \(p^{-}\) is its symmetric counterpart, the annotated keypoint with the opposite object-side label. We use the horizontal-flip keypoint pairs released with GeoAware-SC \cite{r4}. The eligible set \(\mathcal{T}\) contains queries with both parts visible and separated by at least \(0.2\) times the longer side of the target bounding box. An accurate comparison assigns the larger score to the true target, and a tie earns half credit. Using \(h(z) = \mathbf{1}\lbrack z > 0\rbrack + \frac{1}{2}\mathbf{1}\lbrack z = 0\rbrack\), the accuracy over eligible queries is

\begin{equation}
\operatorname{Acc}(G) = \frac{1}{|\mathcal{T}|}\sum_{i\in\mathcal{T}}^{}h\left( s_{G}(p_{i}^{+} \mid q_{i}) - s_{G}(p_{i}^{-} \mid q_{i}) \right).
\label{eq:2}
\end{equation}

The primary feature scores have no exact ties, whereas binary position scores can tie. For coordinate matching, PCK evaluates the predicted location \(\hat{p}\). A SPair query is correct when \(\parallel \hat{p} - p^{+} \parallel_{2} < \alpha \max(w_{t},h_{t})\), using the width and height of the target box. A left/right swap falls within the threshold of \(p^{-}\) and outside it for \(p^{+}\). We count swaps on all 51,972 queries with a visible target counterpart, including nearby parts. Comparing two matching rules, a correction changes an incorrect match to a correct one, and an introduced error changes a correct match to an incorrect one. The net gain is the number of corrections minus the number of introduced errors.

\subsection{Datasets and Experimental Settings}
\label{sec:datasets-and-experimental-settings}

SPair-71k \cite{r5} provides 12,234 test pairs and 88,328 visible queries across 18 categories. We use 5,384 validation pairs with 39,244 queries for calibration and 322 unlabeled validation images to estimate spatial means. SPair-U \cite{r37} provides new keypoint annotations on the SPair-71k test pairs, and our custom AP-10K evaluation \cite{r38} contains 3,879 distinct ordered pairs of cropped animals with 24,234 visible and 4,185 eligible queries. We retain the first occurrence of each ordered pair in the original 4,000-entry test list. For PF-PASCAL \cite{r6}, we use the SCNet test split \cite{r39} of 299 pairs and 2,414 queries. We also use a subset without detected overlap with the previously processed SPair-71k image pool. The audit flags a pair if either image shares a VOC identifier or exact file hash with that pool, or has a 64-bit difference hash within Hamming distance four. The subset contains 162 pairs and 1,285 queries.

We report per-keypoint PCK@0.10, the percentage of correct predictions across all evaluated queries. SPair normalizes errors by the longer side of the target bounding box, and PF-PASCAL uses the longer image side. Both use a strict distance threshold.

In the primary SPair-71k comparisons, DINOv2 and MARCO use images resized to a long side of 840 pixels and placed at the upper left of an 840 \(\times\) 840 canvas, with padding on the right and bottom as needed. The frozen ViT-L/14 with four register tokens uses its final patch tokens on a 60 \(\times\) 60 grid with the adapter scales set to zero. MARCO v1.0 uses the adapted tokens and its released upsampling head on a 240 \(\times\) 240 grid. These extractions reflect the image content before padding, while the architectural symmetrization test reflects the padded canvas. DINOv3 uses the timm ViT-L/16 LVD-1689M checkpoint \cite{r40} at 960 and 864 pixels, and the PF-PASCAL comparison uses MARCO at 840 and 756 pixels. CLIP ViT-L/14 uses patch tokens from its penultimate transformer block (448-pixel canvas, 32 \(\times\) 32 grid). ResNet-50 uses layer 3 (833-pixel canvas, 53 \(\times\) 53 grid). Both use padded-canvas reflection. We fix inference settings before execution and use saved outputs for negation of the odd score and analyses of groups defined by all five position scores.

\subsection{Position Agreement and Conflict}
\label{sec:position-agreement-and-conflict}

We define position agreement and conflict by comparing the two annotated target locations with a binary box-side score. The score compares the source query and each target location relative to their respective bounding-box centers, favoring locations on the same side. In \(\mathcal{T}\), it favors the true part for 14,492 queries, the counterpart for 4,587, and ties for 9,899. Together with the 22,994 queries with a nearby symmetric counterpart and the 36,356 without a visible annotated counterpart, these groups partition all 88,328 visible test queries for the matching analysis. To examine how the comparisons depend on the position score, we use binary box side, binary image side, nearest normalized image \(x\), nearest normalized box \(x\), and image-side product. Let \(x_{s}\) and \(x_{c}\) be the horizontal coordinates of the source query and a target candidate. Let \(W_{s},W_{t}\) denote the content-image widths, \(b_{s},b_{t}\) the box centers, and \(w_{s},w_{t}\) the box widths. Then

\begin{equation}
\left\{ \begin{aligned}
u_{s} & = x_{s}/W_{s} - \frac{1}{2}, & u_{c} & = x_{c}/W_{t} - \frac{1}{2}, \\
v_{s} & = (x_{s} - b_{s})/w_{s}, & v_{c} & = (x_{c} - b_{t})/w_{t}, \\
P_{x} & = - |u_{s} - u_{c}|, & P_{prod} & = 4u_{s}u_{c}.
\end{aligned} \right.\ 
\label{eq:3}
\end{equation}

Nearest normalized box \(x\) uses \(- |v_{s} - v_{c}|\). The binary scores are

\[\left\{ \begin{aligned}
P_{img} & = \mathbf{1}\lbrack u_{s}u_{c} > 0\rbrack, \\
P_{box} & = \mathbf{1}\lbrack v_{s}v_{c} > 0\rbrack.
\end{aligned} \right.\ \]

The image coordinates refer to content before padding, and the box coordinates use annotated bounding boxes. Subtracting one half from both normalized image \(x\) coordinates leaves their distance unchanged. The image-side product multiplies their signed horizontal offsets from their respective image centers. These scores also define a stricter, post hoc conflict subset of 2,417 eligible queries for which all five scores strictly favor the symmetric counterpart over the true target.

The primary comparison holds the target image and the two annotated locations fixed across position conditions. We group queries by category, target image and annotated part pair, then retain groups that contain both agreement and conflict under the binary box-side score. Source queries can differ between conditions. We average within each condition and group, average the groups within each target, and weight targets equally. In the stricter comparison, all five scores must favor the correct target keypoint for agreement or the counterpart for conflict. The four models are evaluated on the same query identities with their respective extraction protocols, including a horizontally reflected copy of the same diffusion-noise tensor for supervised GeoAware-SC (Section 3.7.1).

To describe how consistently the five position scores favor one target part, we keep all 28,978 eligible queries and sum the signs of the five score differences. For position scores \(P_{j}\), the sum is \(r = \sum_{j = 1}^{5}\operatorname{sign}\lbrack P_{j}(p^{+} \mid q) - P_{j}(p^{-} \mid q)\rbrack\). Its eleven values run from unanimous preference for the counterpart (\(- 5\)) to unanimous preference for the correct target keypoint (\(5\)). These groups cover 306 target images. We weight queries equally within each group.

\subsection{Aligned Reflection and Descriptor Decomposition}
\label{sec:aligned-reflection-and-descriptor-decomposition}

\begin{figure}[!htbp]
\centering
\includegraphics[width=0.82\linewidth,height=0.77\textheight,keepaspectratio]{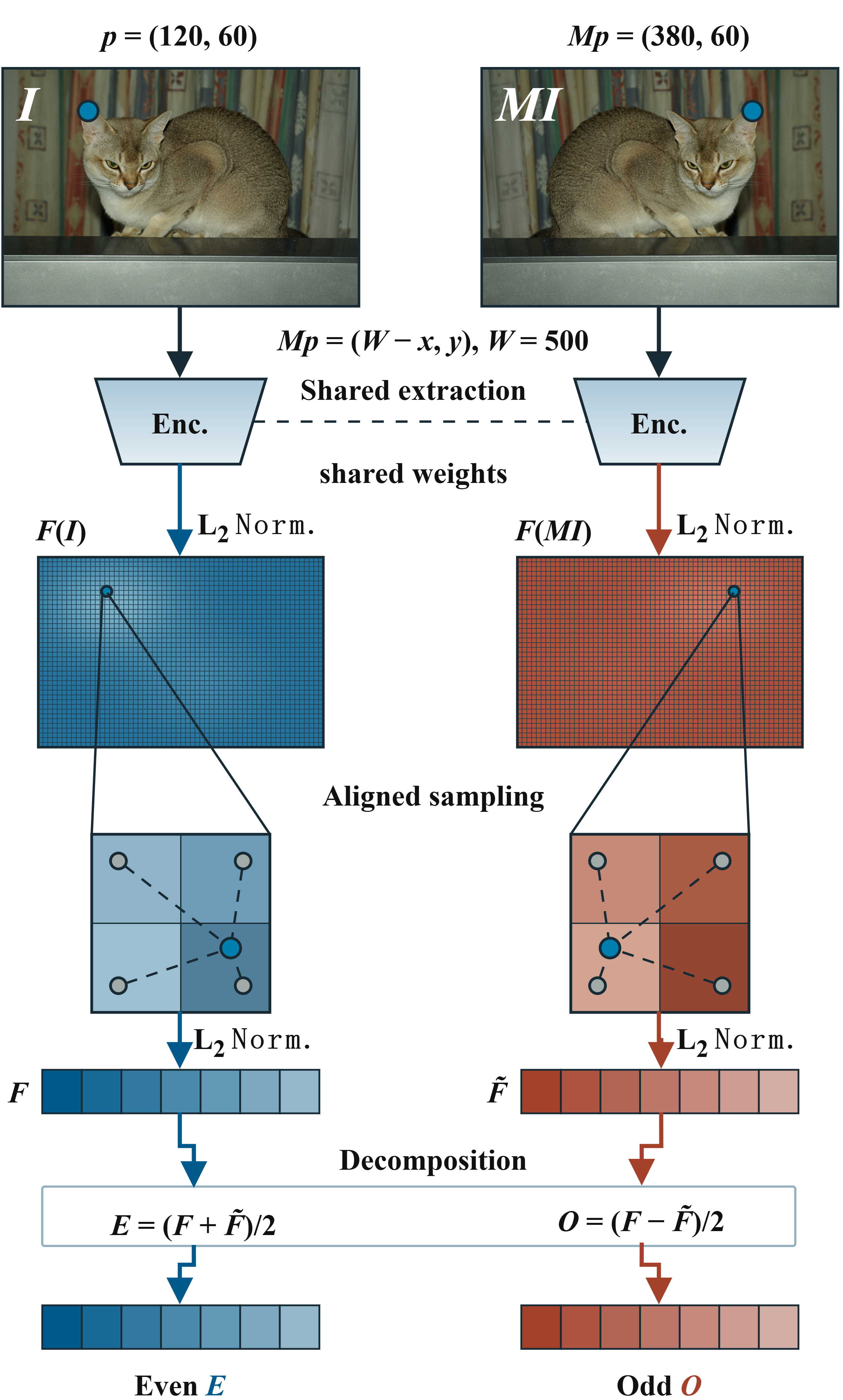}
\caption{Sampling the same content location before and after horizontal reflection. The half-sum gives the even descriptor and the half-difference gives the odd descriptor.}
\label{fig:2}
\end{figure}

Figure~\ref{fig:2} follows an annotated location through an image and its horizontal reflection. Let \(M\) act on the image and its coordinates. The shared extractor produces a feature grid for each image. Sampling at corresponding locations makes the two descriptors refer to the same point in the image content. Their half-sum and half-difference define the even and odd components of the aligned response, respectively. The feature-grid colors in Fig.~\ref{fig:2} are schematic.

\begin{equation}
\left\{ \begin{aligned}
\tilde{F}(I)(p) & = F(MI)(Mp), \\
E & = \frac{1}{2}(F + \tilde{F}), \\
O & = \frac{1}{2}(F - \tilde{F}).
\end{aligned} \right.\ 
\label{eq:4}
\end{equation}

For DINOv2 and MARCO, the same fixed extractor processes \(I\) and \(MI\). We \(L_{2}\)-normalize the grid vectors and normalize again after bilinear sampling at \(p\) and \(Mp\). In the continuous content coordinates illustrated in Fig.~\ref{fig:2}, \(Mp = (W - x,y)\), where \(W\) is the image width. The two unit descriptors enter Eq. (4), and \(E\) and \(O\) retain their resulting norms at this step. Repeating the reflection exchanges the two samples under an involutive convention, which gives the component responses in Eq. (5).

\begin{equation}
\left\{ \begin{aligned}
E(MI)(Mp) & = E(I)(p), \\
O(MI)(Mp) & = - O(I)(p).
\end{aligned} \right.\ 
\label{eq:5}
\end{equation}

For unit \(F,\tilde{F}\), the decomposition gives \(F = E + O\), \(E^{\mathsf{T}}O = 0\), and \(\parallel E \parallel^{2} + \parallel O \parallel^{2} = 1\) in exact arithmetic. We measure odd energy as \(\parallel O \parallel^{2} = (1 - \langle F,\tilde{F}\rangle)/2\) before cosine normalization, and summarize it by the median across visible queries.

\textbf{Proposition 1. Cosine margins under target reflection}

Fix the source image \(I_{s}\) and query location \(q\). For component \(G\), its source descriptor is \(a_{G} = G(I_{s})(q)\). Given labeled target observation \(z = (I_{t},p^{+},p^{-})\), the transformed observation \(Rz = (MI_{t},Mp^{-},Mp^{+})\) reflects the target image and exchanges the part labels. For the expected accuracy below, assume that \(z\) and \(Rz\) have the same conditional distribution after visibility selection. Define the cosine margin \(d_{G}(z) = \cos(a_{G},G(I_{t})(p^{+})) - \cos(a_{G},G(I_{t})(p^{-}))\). For the aligned even component, \(d_{E}(Rz) = - d_{E}(z)\). With \(h(d) = \mathbf{1}\lbrack d > 0\rbrack + \frac{1}{2}\mathbf{1}\lbrack d = 0\rbrack\), each pair satisfies \(h(d) + h( - d) = 1\), so the expected accuracy of the even component is one half. For the raw odd component, both target scores change sign and exchange roles, giving \(d_{O}(Rz) = d_{O}(z)\). The odd component therefore preserves the ordering even when it favors the wrong part. A zero-norm descriptor receives a cosine score of zero, and tied scores receive half credit.

We also measure the odd response relative to its average at each location across validation images. The spatial mean and centered response are defined in Eq. (6).

\begin{equation}
\left\{ \begin{aligned}
\mu(p) & = \frac{1}{|\mathcal{V}_{p}|}\sum_{J \in \mathcal{V}_{p}}^{}O(J)(p), \\
O_{res}(I)(p) & = O(I)(p) - \mu(p),
\end{aligned} \right.\ 
\label{eq:6}
\end{equation}

Here \(\mathcal{V}_{p}\) contains validation images with content at \(p\). Each model and geometry has its own mean, estimated from unlabeled validation images and shared by source and target. We compute the mean from unnormalized odd vectors and subtract it before cosine normalization. MARCO's primary matching rule uses raw \(O\), with centered scores retained as auxiliary measurements. Section 3.5 describes how the validation mean is sampled for the upsampling comparison.

\subsection{Feature Upsampling and Architectural Controls}
\label{sec:feature-upsampling-and-architectural-controls}

We compare four readouts of the same MARCO backbone and adapter output: (1) its native 60 \(\times\) 60 feature grid, (2) bilinear interpolation to 240 \(\times\) 240, (3) the learned 240 \(\times\) 240 output produced by MARCO's upsampling head (two successive 2\(\times\) upsampling blocks, each comprising a transposed convolution, GELU, a depthwise convolution, and channel normalization), and (4) the learned 240 \(\times\) 240 output average-pooled back down to 60 \(\times\) 60. We compare all four readouts under the same 840-pixel input geometry, with the image placed at the upper left of the canvas (Table~\ref{tab:2}). We extract the readouts in a shared forward pass for each image orientation. To compare the learned and native features at the same grid size, we average nonoverlapping 4 \(\times\) 4 blocks of the unnormalized learned output in float32. We then normalize the descriptors and sample the same annotated points in each conflict group as in Section 3.4.

For Fig.~\ref{fig:6}(a--c), the long side of the content is 672 pixels on an 840-pixel canvas, with upper-left offsets of (84, 84) for the source and (0, 84) for the target. We score raw full and odd descriptors on all 28,978 eligible queries, including 4,587 box-side conflicts and 2,417 conflicts under all five position scores. Only analyses that subtract a validation mean require valid mean samples at all three annotated locations. We estimate a separate mean from centrally placed validation images for each readout, positional-embedding condition and reflection convention.

The translation comparison tests sensitivity to absolute image position while preserving the content. We resize the longer side to 672 pixels on an 840-pixel canvas and use upper-left target offsets (0, 84), (14, 84), and (168, 84) pixels, varying the horizontal coordinate with the source fixed at (84, 84). We evaluate reflection before padding and reflection of the padded image separately and retain all eligible queries for raw \(F/O\).

To examine how positional embeddings (PE) affect the reflection response, we horizontally symmetrize the absolute embedding \(B\) as \(B_{sym} = (B + MB)/2\). Two random controls match the Frobenius norm of the embedding change. One uses Gaussian perturbations, and the other uses orthogonal rotations that preserve the squared norm of the odd embedding component. Each uses five prespecified seeds, and we average the per-query accuracy or correctness indicators across interventions. These absolute-embedding perturbations apply to DINOv2 and MARCO. DINOv3 uses rotary positional embeddings (RoPE). A second architectural comparison also symmetrizes patch embedding kernels, both separately and jointly with the absolute embeddings. Before upsampling, self-attention is equivariant to token permutations, and the feed-forward layers, layer normalization and MARCO adapters operate separately on each token. The class and register token indices remain fixed under reflection. This comparison uses padded-canvas reflection and evaluates MARCO's feature grid with adapters enabled, before learned upsampling.

\subsection{Candidate Generation and Selection}
\label{sec:candidate-generation-and-selection}

\begin{figure}[!htbp]
\centering
\includegraphics[width=\linewidth,height=0.77\textheight,keepaspectratio]{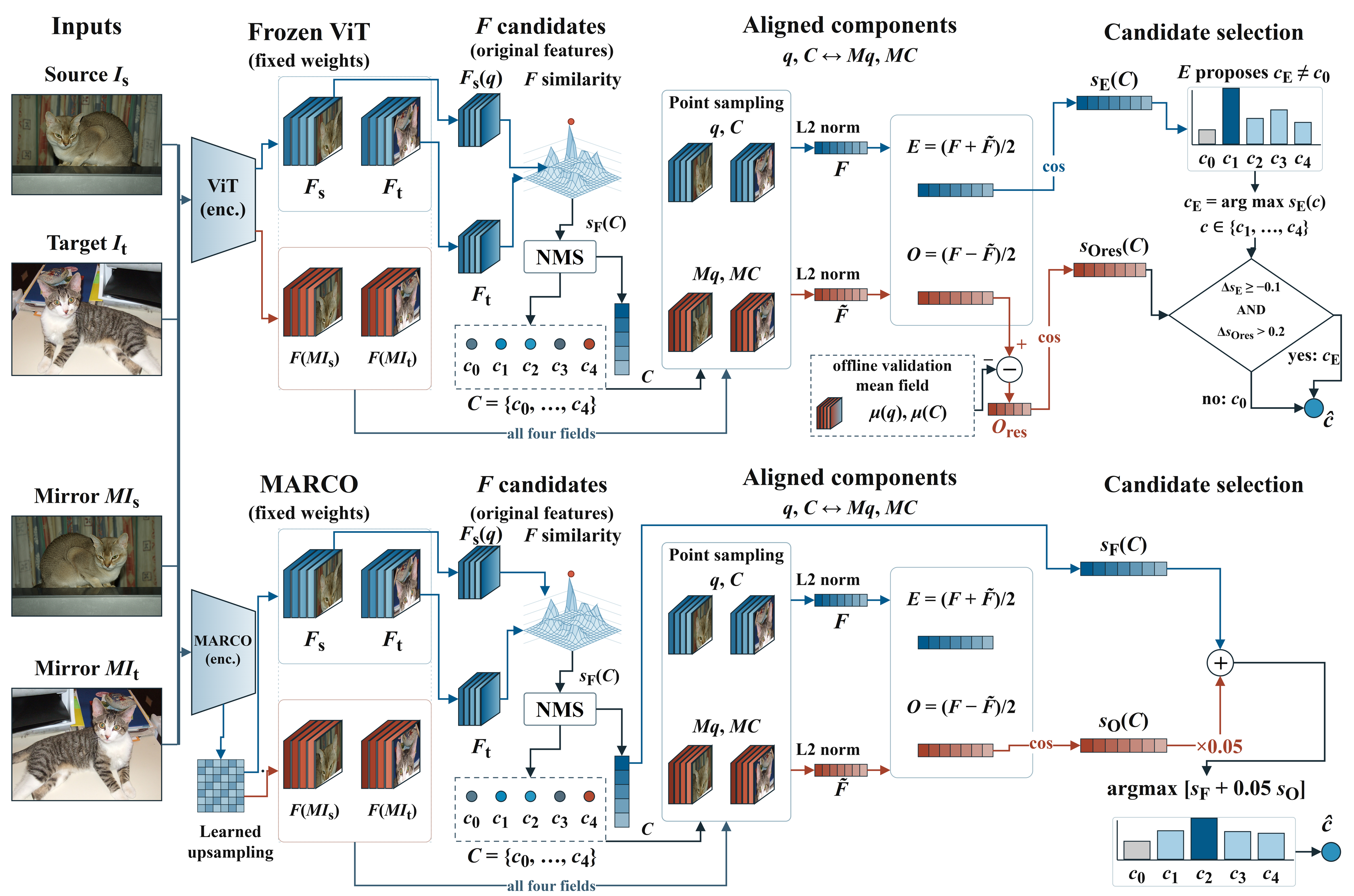}
\caption{Candidate selection for frozen ViTs and MARCO. Each model processes the source and target images and their horizontal reflections. MARCO applies the same learned upsampling head to features from the original and reflected images. Each model selects from its own five peaks in the original similarity map. A frozen ViT accepts the alternative proposed by the even score when both conditions in Eq. (7) hold. MARCO selects the peak with the highest combined original and weighted odd score (Eq. (8)).}
\label{fig:3}
\end{figure}

The candidate rules test whether the mirror components can help choose a coordinate from peaks in the original similarity map (Fig.~\ref{fig:3}). Candidate set \(\mathcal{C}\) contains the original matcher's \(K = 5\) highest similarity peaks after nonmaximum suppression (NMS) with a radius of 42 pixels in the padded geometry. Let \(c_{0}\) be the highest-scoring original peak, which is also the prediction under hard argmax, and \(c_{E}\) the alternative with greatest even similarity among the other four candidates. For frozen DINOv2, we switch to \(c_{E}\) when the even and centered odd margins satisfy both conditions in Eq. (7). Each model uses its own candidates. Figure~\ref{fig:3} uses schematic feature grids and score bars; its score differences are defined by \(\Delta s_{G} = s_{G}(c_{E} \mid q) - s_{G}(c_{0} \mid q)\).

\begin{equation}
\left\{ \begin{aligned}
s_{E}(c_{E}) - s_{E}(c_{0}) & \geq - \delta, \\
s_{O_{res}}(c_{E}) - s_{O_{res}}(c_{0}) & > \tau.
\end{aligned} \right.\ 
\label{eq:7}
\end{equation}

If alternatives tie for the highest even score, we take the first such alternative in candidate order. We keep \(c_{0}\) if either condition in Eq. (7) fails, including equality at the centered odd-score threshold \(\tau\). Validation selects \(\delta = 0.1,\tau = 0.2\), and DINOv3 uses the same values on a 960-pixel canvas. Figure~\ref{fig:3} depicts the 840-pixel canvas used by DINOv2 and MARCO. For MARCO, we add the weighted raw odd score to each candidate's original similarity and choose the highest combined score according to Eq. (8).

\begin{equation}
\left\{ \begin{aligned}
\hat{c} & = \operatorname*{arg\,max}_{c\in\mathcal{C}}\{ s_{F}(c) + \gamma s_{O}(c)\}, \\
\gamma & = 0.05.
\end{aligned} \right.\ 
\label{eq:8}
\end{equation}

With model weights fixed, we select rule parameters on labeled validation pairs and estimate spatial means from unlabeled validation images. At test time, the rules score candidates using source-to-candidate feature similarities, and counterpart annotations and conflict groups define the evaluation strata. In the primary five-peak MARCO comparisons, we report the selected peak coordinate on the upsampled grid. The released MARCO localizer \cite{r3} instead computes a full target dot-product map, weights it with a Gaussian centered on its maximum (standard deviation seven output cells), and returns the expected coordinate under a global softmax after dividing the weighted logits by ten. GeoAware-SC uses MARCO's settings \(K = 5,\gamma = 0.05\) and localizes its chosen candidate as described in Section 3.7.1. The position-based baselines that depend on annotated boxes are described in Section 3.7.

For the additional MARCO comparison, three choices of center shared the same unnormalized learned target field, Gaussian weighting and global softmax. A used the maximum of the full dot-product map from the released source sampler. B used the highest cosine peak on content-valid cells. C selected among the same five cosine peaks with the fixed additive odd rule in Eq. (8). A scored the entire canvas, whereas B and C used the study's content crop, normalized descriptors and sampling convention to select centers. C versus B isolates odd reranking with the decoder fixed. We report per-query PCK and mean per-pair PCK, weighting each nonempty image pair equally in the latter. For contrasts, we resampled physical target images 2,000 times, applying the same draws to both conditions.

\subsection{Inference Baselines and Matching Controls}
\label{sec:inference-baselines-and-matching-controls}

We use the position scores from Section 3.3 as baselines on the same candidate sets. Each score can rank the candidates directly (Position score only) or contribute a weighted term to the original similarity (Original + position score). For frozen ViTs, the Even-score candidate with position test baseline substitutes the position score for the centered odd score in Eq. (7), while retaining the even-score proposal and tolerance. Ties between candidates preserve the original order. We select settings by validation PCK separately for baselines with and without annotated boxes, preferring smaller weights or thresholds if validation scores tie. DINOv3 uses the DINOv2 settings. Binary box side and nearest normalized box x require annotated boxes; the three image-based scores do not.

The other matching baselines change feature aggregation or alignment. Horizontal-flip test-time augmentation averages aligned descriptors from the original and flipped images. Two-scale inference averages cosine scores at 840 and 756 pixels for DINOv2 and MARCO and at 960 and 864 pixels for DINOv3. For pose alignment, we compare the original and horizontally reflected source against the original target. We use the reflected source only if its mean Euclidean distance over mutual nearest-neighbor (MNN) descriptor pairs is smaller. Pose is scored on a padded 60 \(\times\) 60 grid using the same feature pipeline as matching. For MARCO, we downsample and normalize the learned features for pose scoring and use the 240 \(\times\) 240 features for matching. The geometric variant preserves source query identity. The keypoint-ID variant instead reindexes reflected-source predictions by left-right partner labels supplied for the mutually visible query set. That set uses visibility annotations from both images. If the partner is absent, the prediction reverts to the original match; target coordinates are not used to select the orientation or reindex predictions.

\subsubsection{Supervised GeoAware-SC and Window Soft Argmax}
\label{sec:supervised-geoaware-sc-and-window-soft-argmax}

The released supervised GeoAware-SC checkpoint combines DINOv2-B/14 and Stable Diffusion 1.5 features \cite{r41} extracted through ODISE \cite{r42}. CLIP conditions the ODISE captioner, and learned aggregation produces 768-channel features. We run window soft argmax without pose alignment or masks. Candidate selection chooses a peak as the window center; soft argmax then predicts the final coordinate from the similarity map. The square window extends seven cells in each direction on the 60 \(\times\) 60 grid, and localization uses the original similarity map with temperature 0.02. We set scores outside the window to zero before the global softmax. We compute the softmax-weighted coordinate over the entire grid, so out-of-window locations still contribute. For the mirrored image, we compare horizontally reflected shared diffusion noise and the unchanged noise tensor. The error breakdown uses reflected noise.

For the geometry-aware evaluation, a query is eligible when the target image shows at least one additional member of the same semantic subgroup in GeoAware-SC \cite{r4} as the true target keypoint, such as the group containing all four paws. This criterion admits 52,671 queries. The bilateral-swap evaluation uses the 51,972 queries with a visible bilateral counterpart.

\subsubsection{Direction and Magnitude of the Odd-Score Margin}
\label{sec:direction-and-magnitude-of-the-odd-score-margin}

Negating the odd score tests whether matching changes depend on the direction of the odd preference. In this post hoc comparison, we negate the odd score and keep the candidates, even-score proposal and even tolerance fixed. Frozen models switch when the centered odd margin is below \(- \tau\), and MARCO uses \(s_{F} - 0.05s_{O}\). We compare the two signs on identical queries and count all four correctness transitions. On the strict conflict subset, we also calculate the expected PCK from switching a uniformly random subset of the even-score proposals. We match the number of switches made by the positive rule, first among all proposals that pass the even tolerance and then among those with an absolute centered odd margin above 0.2. Each of 2,000 paired bootstrap samples resamples physical targets and recomputes the switching fraction.

\subsection{Additional Datasets and Inference Cost}
\label{sec:additional-datasets-and-inference-cost}

The additional matching comparisons use the custom AP-10K pairs and the additional query annotations in SPair-U. AP-10K uses rule parameters selected on validation. A zero change records a validation choice to keep the original prediction. On SPair-U, we apply the fixed matching rules to the additional annotations.

We measure median inference cost on 108 validation pairs per model and method on a T4 with two CPU threads, excluding two warm-up pairs. End-to-end latency covers decoding through synchronized matching with the model loaded and without cached features. Matching with cached features measures matching alone. Initialization and validation-mean estimation are excluded. The peak allocated GPU memory column reports the median of the 108 per-pair peak allocations. A dash in the GFLOP column denotes a configuration without a separate operator-count estimate. The GFLOP estimates count dense, attention and convolution operations for one image pair and one source query. On validation, we also replay an exploratory rule that requests the reflected pass for queries in the lowest quarter of the gap between the two highest original peaks after NMS. Ties at the threshold are included. A pair requires the extra pass whenever at least one query qualifies.

\section{Results}
\label{sec:results}

\subsection{Discrimination Across Position Groups}
\label{sec:discrimination-across-position-groups}

The raw odd descriptor improves overall left/right discrimination in the frozen models, but the ordering changes when the box-side score favors the symmetric counterpart. On SPair-71k, DINOv2 reaches 70.66\% accuracy with \(O\) and 58.57\% with \(F\) across all eligible queries. In box-side conflicts, the corresponding accuracies fall to 17.68\% and 33.64\%. DINOv3 follows the same pattern, with raw odd accuracy of 77.33\% overall and 27.86\% in conflicts, compared with 36.12\% for its full descriptor in conflicts. Subtracting the average odd response at each location raises conflict accuracy to 49.20\% for DINOv2 and 47.96\% for DINOv3, both close to the 50\% chance level. The comparisons in Table~\ref{tab:1} use the same two annotated target locations for the feature and position scores. The Box-side tie column includes queries with no preference under the box-side score; the final column includes queries where all five position scores favor the counterpart.

Similar overall accuracies can come from different behavior on the same position groups. On the 9,899 queries where the box-side score ties, raw odd accuracy is 70.25\% for DINOv2 and 75.70\% for DINOv3, while the best of the five position scores reaches 60.06\%. For DINOv2, nearest normalized box \(x\) reaches 70.26\% overall, close to the 70.66\% for the odd score. It performs better on non-tied queries (75.56\% versus 70.87\%) and worse on tied queries (60.06\% versus 70.25\%).

MARCO and supervised GeoAware-SC retain higher discrimination when position favors the counterpart. At the original 840-pixel input size, MARCO's raw odd accuracy reaches 89.47\% in box-side conflicts, and its advantage over the full descriptor is 6.08 pp when all five position scores favor the counterpart. GeoAware-SC reaches 83.48\% with the odd descriptor and 81.62\% with the full descriptor in box-side conflicts (Table~\ref{tab:1}). In the separate MARCO setting with content resized to 672 pixels on an 840-pixel canvas, MARCO's odd and full descriptors reach 89.01\% and 86.05\% in box-side conflicts, a paired difference of 2.96 points. CLIP and ResNet-50 also lose odd accuracy under box-side conflict, reaching 39.94\% and 45.61\%, respectively. ResNet-50 exhibits this dependence without an explicit ViT positional embedding.

\begin{table}[!htbp]
\centering
\small
\setlength{\parskip}{0pt}
\setlength{\tabcolsep}{4pt}
\renewcommand{\arraystretch}{1.12}
\caption{Left-Right Discrimination Accuracy of Feature and Position Scores (\%)}\label{tab:1}
\begin{tabular}{@{}
  >{\centering\arraybackslash}p{(\linewidth - 10\tabcolsep) * \real{0.3125}}
  >{\centering\arraybackslash}p{(\linewidth - 10\tabcolsep) * \real{0.1054}}
  >{\centering\arraybackslash}p{(\linewidth - 10\tabcolsep) * \real{0.1455}}
  >{\centering\arraybackslash}p{(\linewidth - 10\tabcolsep) * \real{0.1455}}
  >{\centering\arraybackslash}p{(\linewidth - 10\tabcolsep) * \real{0.1455}}
  >{\centering\arraybackslash}p{(\linewidth - 10\tabcolsep) * \real{0.1455}}@{}}
\toprule\noalign{}
\begin{minipage}[b]{\linewidth}\centering
Score
\end{minipage} & \begin{minipage}[b]{\linewidth}\centering
All
\end{minipage} & \begin{minipage}[b]{\linewidth}\centering
Box-side conflict
\end{minipage} & \begin{minipage}[b]{\linewidth}\centering
Box-side

tie
\end{minipage} & \begin{minipage}[b]{\linewidth}\centering
No box-side tie
\end{minipage} & \begin{minipage}[b]{\linewidth}\centering
All five favor counterpart
\end{minipage} \\
\midrule\noalign{}
DINOv2 \(F\) & 58.57 & 33.64 & 53.32 & 61.30 & 21.27 \\
DINOv2 \(O\) & 70.66 & 17.68 & 70.25 & 70.87 & 2.98 \\
DINOv2 \(O_{res}\) & 70.26 & 49.20 & 72.08 & 69.32 & 47.08 \\
DINOv3 \(F\) & 64.90 & 36.12 & 57.08 & 68.96 & 25.57 \\
DINOv3 \(O\) & 77.33 & 27.86 & 75.70 & 78.17 & 13.07 \\
DINOv3 \(O_{res}\) & 80.48 & 47.96 & 79.24 & 81.12 & 44.48 \\
MARCO \(F\) & 91.35 & 84.63 & 89.04 & 92.55 & 83.66 \\
MARCO \(O\) & 95.36 & 89.47 & 94.15 & 95.98 & 89.74 \\
MARCO \(O_{res}\) & 94.84 & 89.91 & 94.20 & 95.18 & 90.28 \\
GeoAware-SC \(F\) & 90.58 & 81.62 & 88.40 & 91.71 & 80.02 \\
GeoAware-SC \(O\) & 92.24 & 83.48 & 89.86 & 93.47 & 81.96 \\
CLIP \(F\) & 50.01 & 50.14 & 49.91 & 50.06 & 50.72 \\
CLIP \(E\) & 49.94 & 50.29 & 49.68 & 50.08 & 50.43 \\
CLIP \(O\) & 52.30 & 39.94 & 52.78 & 52.05 & 34.59 \\
ResNet-50 \(F\) & 53.66 & 49.88 & 51.38 & 54.85 & 47.87 \\
ResNet-50 \(E\) & 49.80 & 50.12 & 49.25 & 50.08 & 49.28 \\
ResNet-50 \(O\) & 73.19 & 45.61 & 69.37 & 75.18 & 42.90 \\
Binary box side & 67.09 & 0.00 & 50.00 & 75.96 & 0.00 \\
Binary image side & 60.79 & 24.35 & 51.13 & 65.80 & 0.00 \\
Nearest normalized image x & 63.85 & 22.95 & 56.09 & 67.88 & 0.00 \\
Nearest normalized box x & 70.26 & 9.85 & 60.06 & 75.56 & 0.00 \\
Image-side product & 65.45 & 15.70 & 57.77 & 69.43 & 0.00 \\
\bottomrule
\end{tabular}
\end{table}

\begin{figure}[!htbp]
\centering
\includegraphics[width=\linewidth,height=0.77\textheight,keepaspectratio]{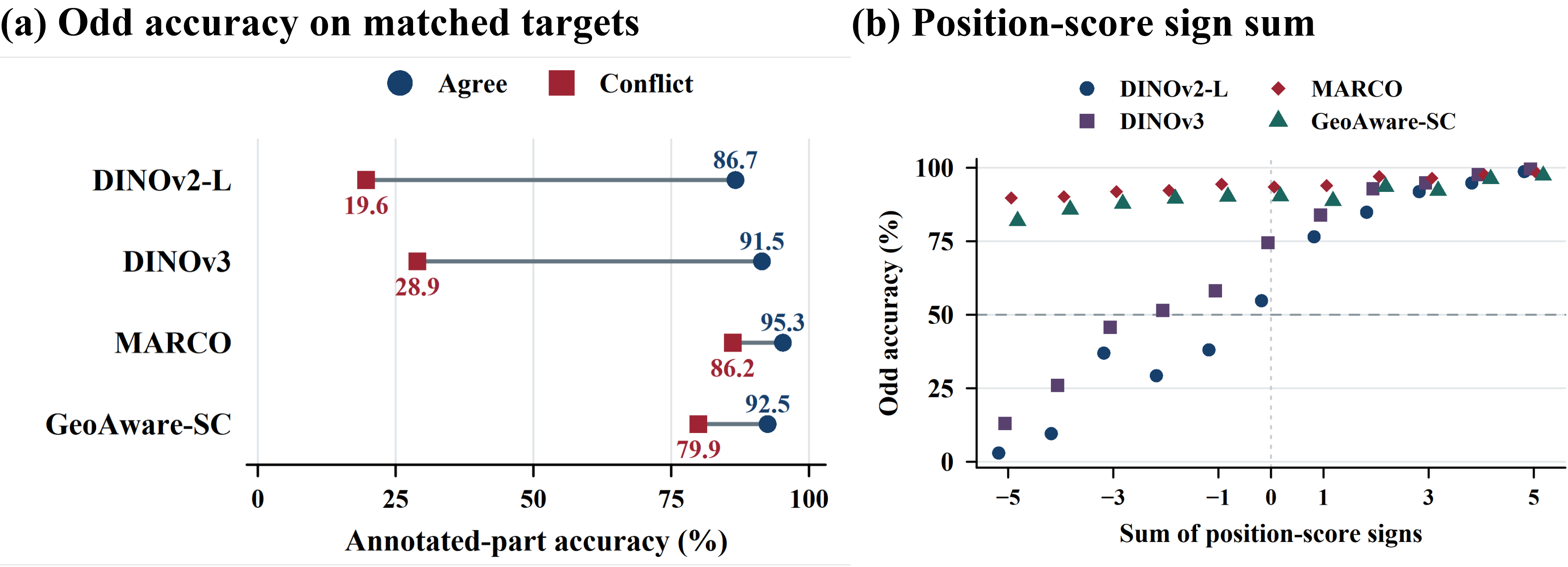}
\caption{Odd discrimination in matched target images and annotated part pairs (a), and across position-score preferences on all eligible queries (b).}
\label{fig:4}
\end{figure}

\begin{figure}[!htbp]
\centering
\includegraphics[width=\linewidth,height=0.77\textheight,keepaspectratio]{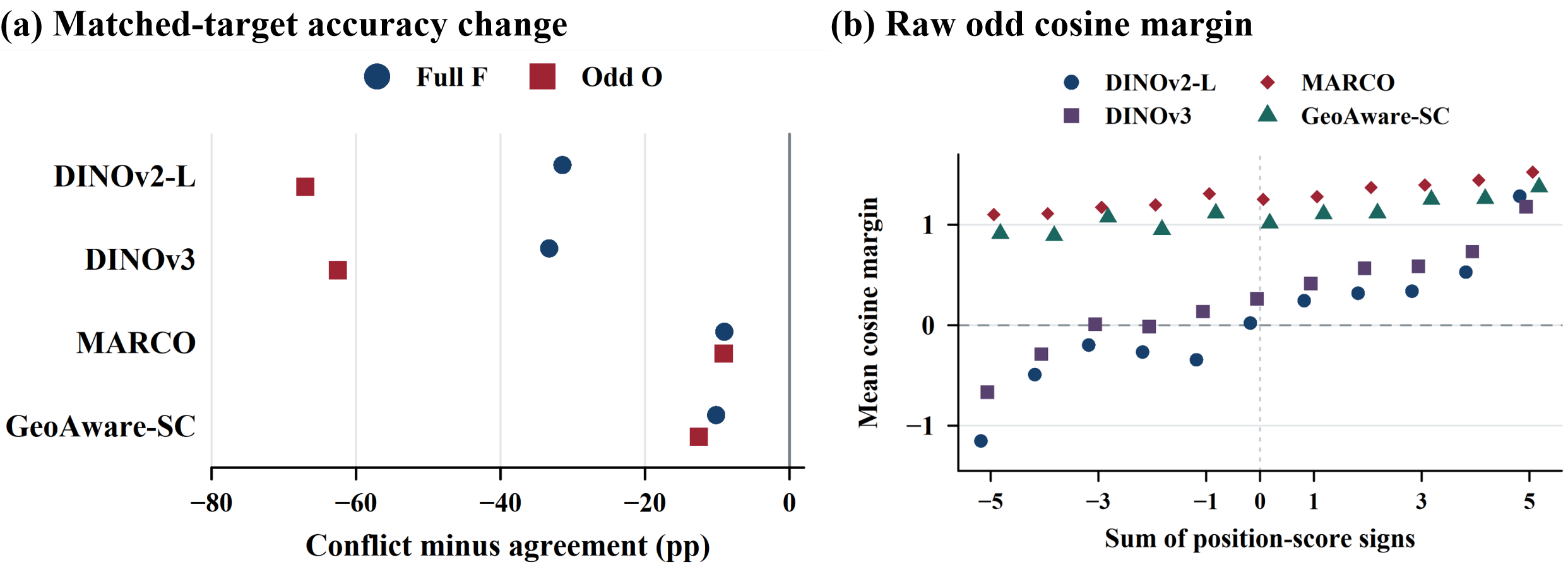}
\caption{Differences in the decline from agreement to conflict (a) and mean odd similarity margins across groups defined by the five position preferences (b).}
\label{fig:5}
\end{figure}

The larger decline in odd accuracy for the frozen models persists when agreement and conflict share the same target image and annotated part pair (Fig.~\ref{fig:4}(a)). This comparison uses the same query identities across models, with 796 groups across 185 targets, 17 categories and 442 source images, comprising 9,847 agreement and 4,556 conflict queries. Raw odd accuracy falls from 86.66\% to 19.63\% for DINOv2 and from 95.27\% to 86.16\% for MARCO. With equal weight per target image, DINOv2's decline exceeds MARCO's by 57.93 pp for odd features and 22.40 points for full features. Thus, the between-model gap in the conflict-related decline is 35.52 points larger for odd than for full descriptors (Fig.~\ref{fig:5}(a)). DINOv3 and supervised GeoAware-SC show odd-accuracy declines of 62.52 and 12.55 points, respectively. In both frozen models the odd component declines more than the full descriptor.

The difference between DINOv2 and MARCO grows under the stricter definition of position conflict. Requiring unanimous agreement or conflict among the five position scores leaves 542 groups on 135 targets, with 5,252 agreement and 2,325 conflict queries. DINOv2's odd-accuracy decline then exceeds MARCO's by 86.82 pp. On the primary matched cohort, the gap is 61.64 percentage points when each query receives equal weight. DINOv2 shows the larger decline whichever category we omit.

Could the matched contrast reflect differences between viewpoint groups? We repeated the comparison within each of the three SPair-71k viewpoint-variation bins, retaining only target images and directed part pairs represented under both position conditions in the same bin. The bins (0, 1 and 2) contained 176, 133 and 30 matched targets, respectively. DINOv2 odd accuracy fell from 87.78\% to 23.11\%, from 84.89\% to 17.07\% and from 76.11\% to 20.28\% across the three bins. MARCO changed from 96.06\% to 84.17\%, from 92.55\% to 91.30\% and from 85.28\% to 81.11\%, respectively. Position agreement distinguished DINOv2 outcomes within every annotated viewpoint bin. Source pose and appearance can still vary within each bin.

Across all 28,978 eligible queries, the frozen models' odd accuracy and mean cosine margins vary substantially with the sum of the five preference signs (Figs. 4(b) and 5(b)). DINOv2 accuracy drops from 36.95\% at \(r = - 3\) to 29.27\% at \(r = - 2\) despite the less negative sum of preference signs. MARCO reaches 89.74\% and 98.12\% in the two extreme groups, where all five scores favor the counterpart and the correct target keypoint, respectively. These panels weight queries equally within each group defined by the sum of the five preference signs and cover the full eligible set. The matched comparison gives equal weight to targets with both conditions.

\subsection{Feature Upsampling and Image Translation}
\label{sec:feature-upsampling-and-image-translation}

\begin{figure}[!htbp]
\centering
\includegraphics[width=\linewidth,height=0.77\textheight,keepaspectratio]{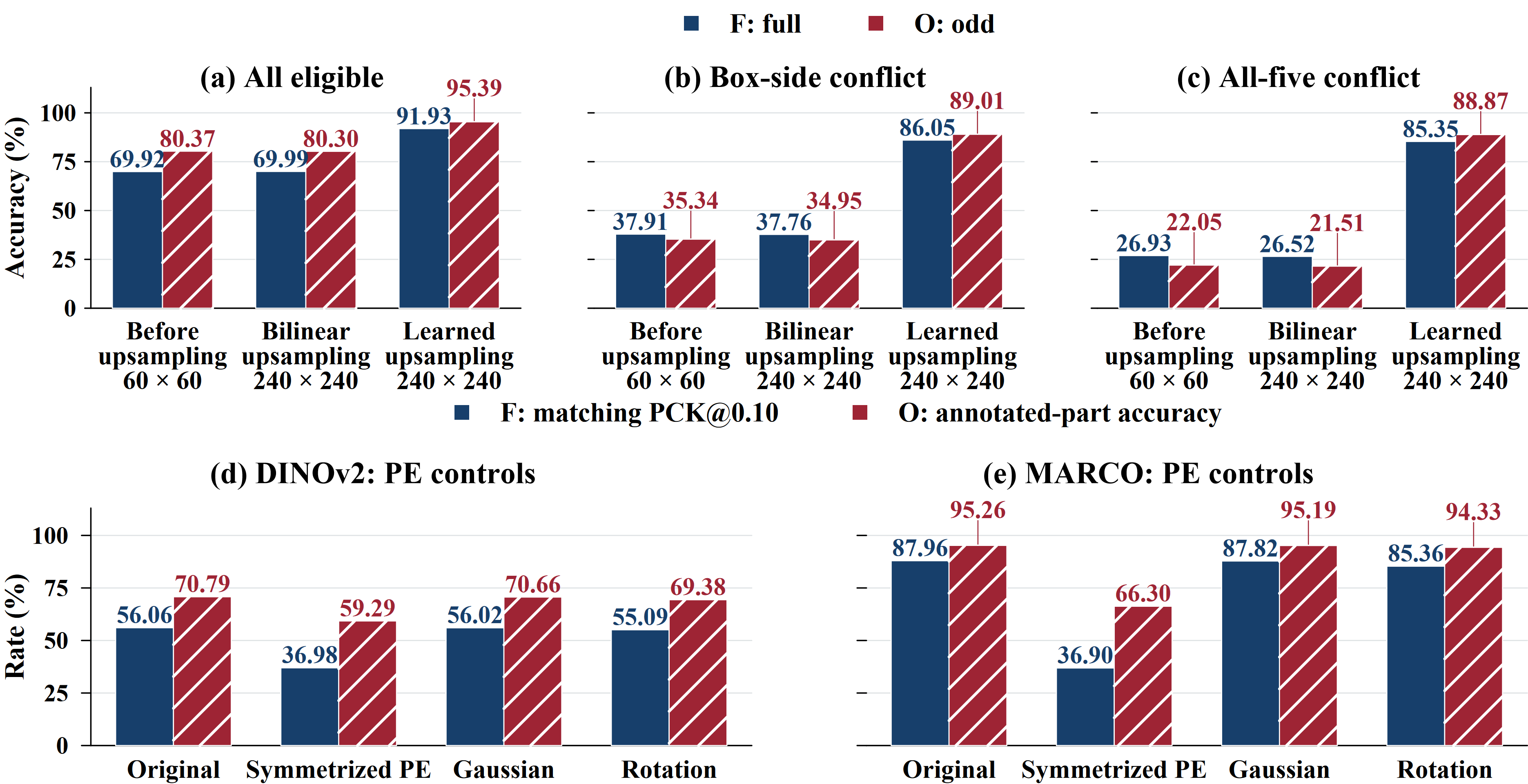}
\caption{MARCO readouts on 28,978 eligible queries, including 4,587 box-side conflicts and 2,417 five-score conflicts (a--c). Positional-embedding interventions change odd discrimination and full-descriptor PCK (d--e).}
\label{fig:6}
\end{figure}

Within MARCO, the choice of feature map makes a large difference to discrimination under position conflict. With the backbone and adapters fixed, native and bilinearly upsampled features have similar accuracy at the annotated parts. Learned upsampling raises odd accuracy by 15.09 pp overall relative to bilinear interpolation. Accuracy increases from 34.95\% to 89.01\% in box-side conflicts and from 21.51\% to 88.87\% when all five position scores favor the counterpart. Full-descriptor accuracy also rises in box-side conflicts, from 37.76\% to 86.05\% (Fig.~\ref{fig:6}(a)-(c)). Translating the same content within the canvas changed odd accuracy by less than half a percentage point for both models, whether we reflected the image before or after padding.

Table~\ref{tab:2} compares four readouts from the same adapted 60 \(\times\) 60 MARCO feature map at the standard input geometry. The learned head takes this map as its only input. On box-side conflicts, odd accuracy was 34.12\% on the native grid and 33.86\% after bilinear interpolation to 240 \(\times\) 240. The learned head reached 89.47\% at 240 \(\times\) 240 and retained 88.84\% after pooling back to 60 \(\times\) 60. On five-score conflicts, the pooled readout likewise exceeded the native readout by 69.67 points. The learned transformation thus made part identity easier to distinguish by cosine similarity even at the original grid size.

\begin{table}[!htbp]
\centering
\small
\setlength{\parskip}{0pt}
\setlength{\tabcolsep}{4pt}
\renewcommand{\arraystretch}{1.12}
\caption{Full and Odd Discrimination Accuracy (\%) at the Standard MARCO Input Geometry}\label{tab:2}
\begin{tabular}{@{}
  >{\centering\arraybackslash}p{(\linewidth - 6\tabcolsep) * \real{0.3566}}
  >{\centering\arraybackslash}p{(\linewidth - 6\tabcolsep) * \real{0.2112}}
  >{\centering\arraybackslash}p{(\linewidth - 6\tabcolsep) * \real{0.2112}}
  >{\centering\arraybackslash}p{(\linewidth - 6\tabcolsep) * \real{0.2112}}@{}}
\toprule\noalign{}
\begin{minipage}[b]{\linewidth}\centering
\textbf{Readout}
\end{minipage} & \begin{minipage}[b]{\linewidth}\centering
\textbf{All eligible; F / O}
\end{minipage} & \begin{minipage}[b]{\linewidth}\centering
\textbf{Box-side conflict; F / O}
\end{minipage} & \begin{minipage}[b]{\linewidth}\centering
\textbf{Five-score conflict; F / O}
\end{minipage} \\
\midrule\noalign{}
Native, 60 \(\times\) 60 & 69.59 / 80.06 & 36.08 / 34.12 & 22.84 / 19.07 \\
Bilinear, 240 \(\times\) 240 & 69.71 / 80.02 & 35.71 / 33.86 & 22.30 / 18.78 \\
Learned, 240 \(\times\) 240 & 91.35 / 95.36 & 84.63 / 89.47 & 83.66 / 89.74 \\
Learned, pooled to 60 \(\times\) 60 & 91.95 / 95.20 & 84.41 / 88.84 & 83.82 / 88.75 \\
\bottomrule
\end{tabular}
\end{table}

Symmetrizing the absolute positional embeddings reduces both odd discrimination and full-descriptor matching PCK. In MARCO, odd accuracy falls from 95.26\% to 66.30\% and PCK from 87.96\% to 36.90\% (Fig.~\ref{fig:6}(d)-(e)). For DINOv2, odd discrimination accuracy starts at 70.79\% in this centered setting; Figure~\ref{fig:6}(d) reports its changes in odd discrimination and full-descriptor PCK separately. Both random controls match the embedding change in Frobenius norm and yield discrimination accuracy and PCK closer to their respective baselines.

The patch kernels also matter for matching. On SPair-71k, symmetrizing either the positional embeddings or the patch kernels separately lowers PCK for both DINOv2-L and MARCO's features before learned upsampling. With both symmetrized, full-descriptor discrimination approaches 50\% for DINOv2-L and MARCO's features before learned upsampling (50.74\% and 50.75\%), and their PCK drops to 10.81\% and 10.87\%, respectively. In this extraction, original DINOv2-L discrimination is 58.54\%, close to the 58.57\% in Table~\ref{tab:1}. Joint symmetrization produces the same pattern on the custom AP-10K pairs (Table~\ref{tab:3}).

\begin{table}[!htbp]
\centering
\scriptsize
\setlength{\parskip}{0pt}
\setlength{\tabcolsep}{4pt}
\renewcommand{\arraystretch}{1.12}
\caption{Effects of Symmetrizing Positional Embeddings and Patch Kernels: Full-Descriptor Discrimination Accuracy and PCK (\%)}\label{tab:3}
\begin{tabular}{@{}
  >{\centering\arraybackslash}p{(\linewidth - 18\tabcolsep) * \real{0.1406}}
  >{\centering\arraybackslash}p{(\linewidth - 18\tabcolsep) * \real{0.1563}}
  >{\centering\arraybackslash}p{(\linewidth - 18\tabcolsep) * \real{0.0879}}
  >{\centering\arraybackslash}p{(\linewidth - 18\tabcolsep) * \real{0.0879}}
  >{\centering\arraybackslash}p{(\linewidth - 18\tabcolsep) * \real{0.0879}}
  >{\centering\arraybackslash}p{(\linewidth - 18\tabcolsep) * \real{0.0879}}
  >{\centering\arraybackslash}p{(\linewidth - 18\tabcolsep) * \real{0.0879}}
  >{\centering\arraybackslash}p{(\linewidth - 18\tabcolsep) * \real{0.0879}}
  >{\centering\arraybackslash}p{(\linewidth - 18\tabcolsep) * \real{0.0879}}
  >{\centering\arraybackslash}p{(\linewidth - 18\tabcolsep) * \real{0.0880}}@{}}
\toprule\noalign{}
\multirow{2}{=}{\centering\arraybackslash \begin{minipage}[b]{\linewidth}\centering
Dataset
\end{minipage}} & \multirow{2}{=}{\centering\arraybackslash \begin{minipage}[b]{\linewidth}\centering
Model
\end{minipage}} & \multicolumn{2}{>{\centering\arraybackslash}p{(\linewidth - 18\tabcolsep) * \real{0.1757} + 2\tabcolsep}}{%
\begin{minipage}[b]{\linewidth}\centering
Original
\end{minipage}} & \multicolumn{2}{>{\centering\arraybackslash}p{(\linewidth - 18\tabcolsep) * \real{0.1757} + 2\tabcolsep}}{%
\begin{minipage}[b]{\linewidth}\centering
Symmetrized PE
\end{minipage}} & \multicolumn{2}{>{\centering\arraybackslash}p{(\linewidth - 18\tabcolsep) * \real{0.1757} + 2\tabcolsep}}{%
\begin{minipage}[b]{\linewidth}\centering
Symmetrized patch kernels
\end{minipage}} & \multicolumn{2}{>{\centering\arraybackslash}p{(\linewidth - 18\tabcolsep) * \real{0.1759} + 2\tabcolsep}@{}}{%
\begin{minipage}[b]{\linewidth}\centering
Both symmetrized
\end{minipage}} \\
& & \begin{minipage}[b]{\linewidth}\centering
Acc.
\end{minipage} & \begin{minipage}[b]{\linewidth}\centering
PCK
\end{minipage} & \begin{minipage}[b]{\linewidth}\centering
Acc.
\end{minipage} & \begin{minipage}[b]{\linewidth}\centering
PCK
\end{minipage} & \begin{minipage}[b]{\linewidth}\centering
Acc.
\end{minipage} & \begin{minipage}[b]{\linewidth}\centering
PCK
\end{minipage} & \begin{minipage}[b]{\linewidth}\centering
Acc.
\end{minipage} & \begin{minipage}[b]{\linewidth}\centering
PCK
\end{minipage} \\
\midrule\noalign{}
SPair-71k & DINOv2-L & 58.54 & 57.04 & 51.07 & 35.75 & 57.17 & 30.29 & 50.74 & 10.81 \\
SPair-71k & MARCO (before upsampling) & 69.59 & 71.18 & 53.13 & 30.41 & 62.87 & 30.99 & 50.75 & 10.87 \\
AP-10K & DINOv2-L & 55.68 & 44.45 & 50.35 & 22.02 & 53.88 & 17.99 & 51.21 & 7.99 \\
AP-10K & MARCO (before upsampling) & 60.81 & 48.11 & 50.70 & 16.85 & 56.51 & 13.94 & 50.90 & 8.41 \\
\bottomrule
\end{tabular}
\end{table}

\subsection{Discrimination Across Backbones and Datasets}
\label{sec:discrimination-across-backbones-and-datasets}

The odd component also improves on the full descriptor in CLIP and ResNet-50, with gains that vary by model and dataset. On SPair-71k, odd scores improve on the full descriptor by 2.3 pp for CLIP and 19.5 points for ResNet-50. On the custom AP-10K pairs, ResNet-50 gains 4.0 points, whereas CLIP changes by 0.5 points. Across all the model and dataset combinations in Table~\ref{tab:4}, even descriptors perform close to 50\% at the two annotated locations. Centered odd scores were not evaluated for these two models; dashes mark these entries in the table.

\begin{table}[!htbp]
\centering
\small
\setlength{\parskip}{0pt}
\setlength{\tabcolsep}{4pt}
\renewcommand{\arraystretch}{1.12}
\caption{Left-Right Discrimination Accuracy and Median Odd Energy}\label{tab:4}
\begin{tabular}{@{}
  >{\centering\arraybackslash}p{(\linewidth - 14\tabcolsep) * \real{0.1489}}
  >{\centering\arraybackslash}p{(\linewidth - 14\tabcolsep) * \real{0.2138}}
  >{\centering\arraybackslash}p{(\linewidth - 14\tabcolsep) * \real{0.0762}}
  >{\centering\arraybackslash}p{(\linewidth - 14\tabcolsep) * \real{0.0762}}
  >{\centering\arraybackslash}p{(\linewidth - 14\tabcolsep) * \real{0.0764}}
  >{\centering\arraybackslash}p{(\linewidth - 14\tabcolsep) * \real{0.0764}}
  >{\centering\arraybackslash}p{(\linewidth - 14\tabcolsep) * \real{0.1526}}
  >{\centering\arraybackslash}p{(\linewidth - 14\tabcolsep) * \real{0.1794}}@{}}
\toprule\noalign{}
\begin{minipage}[b]{\linewidth}\centering
Dataset
\end{minipage} & \begin{minipage}[b]{\linewidth}\centering
Model
\end{minipage} & \begin{minipage}[b]{\linewidth}\centering
\(F\) (\%)
\end{minipage} & \begin{minipage}[b]{\linewidth}\centering
\[E\]

(\%)
\end{minipage} & \begin{minipage}[b]{\linewidth}\centering
\[O\]

(\%)
\end{minipage} & \begin{minipage}[b]{\linewidth}\centering
\[O_{res}\]

(\%)
\end{minipage} & \begin{minipage}[b]{\linewidth}\centering
Median odd energy

(\%)
\end{minipage} & \begin{minipage}[b]{\linewidth}\centering
\(O - F\) accuracy (pp)
\end{minipage} \\
\midrule\noalign{}
SPair-71k & MARCO & 91.4 & 50.1 & 95.4 & 94.8 & 3.9 & +4.0 \\
SPair-71k & DINOv2-L & 58.6 & 49.9 & 70.7 & 70.3 & 4.8 & +12.1 \\
AP-10K & MARCO & 72.9 & 50.9 & 84.3 & 82.3 & 3.1 & +11.5 \\
AP-10K & DINOv2-L & 55.7 & 50.1 & 62.1 & 67.7 & 4.4 & +6.4 \\
SPair-71k & CLIP ViT-L/14 & 50.0 & 49.9 & 52.3 & - & 4.1 & +2.3 \\
SPair-71k & ResNet-50 & 53.7 & 49.8 & 73.2 & - & 4.6 & +19.5 \\
AP-10K & CLIP ViT-L/14 & 50.4 & 49.9 & 50.8 & - & 2.2 & +0.5 \\
AP-10K & ResNet-50 & 51.1 & 50.7 & 55.1 & - & 4.2 & +4.0 \\
\bottomrule
\end{tabular}
\end{table}

\subsection{Matching Against Position-Based Baselines}
\label{sec:matching-against-position-based-baselines}

\begin{figure}[!htbp]
\centering
\includegraphics[width=\linewidth,height=0.77\textheight,keepaspectratio]{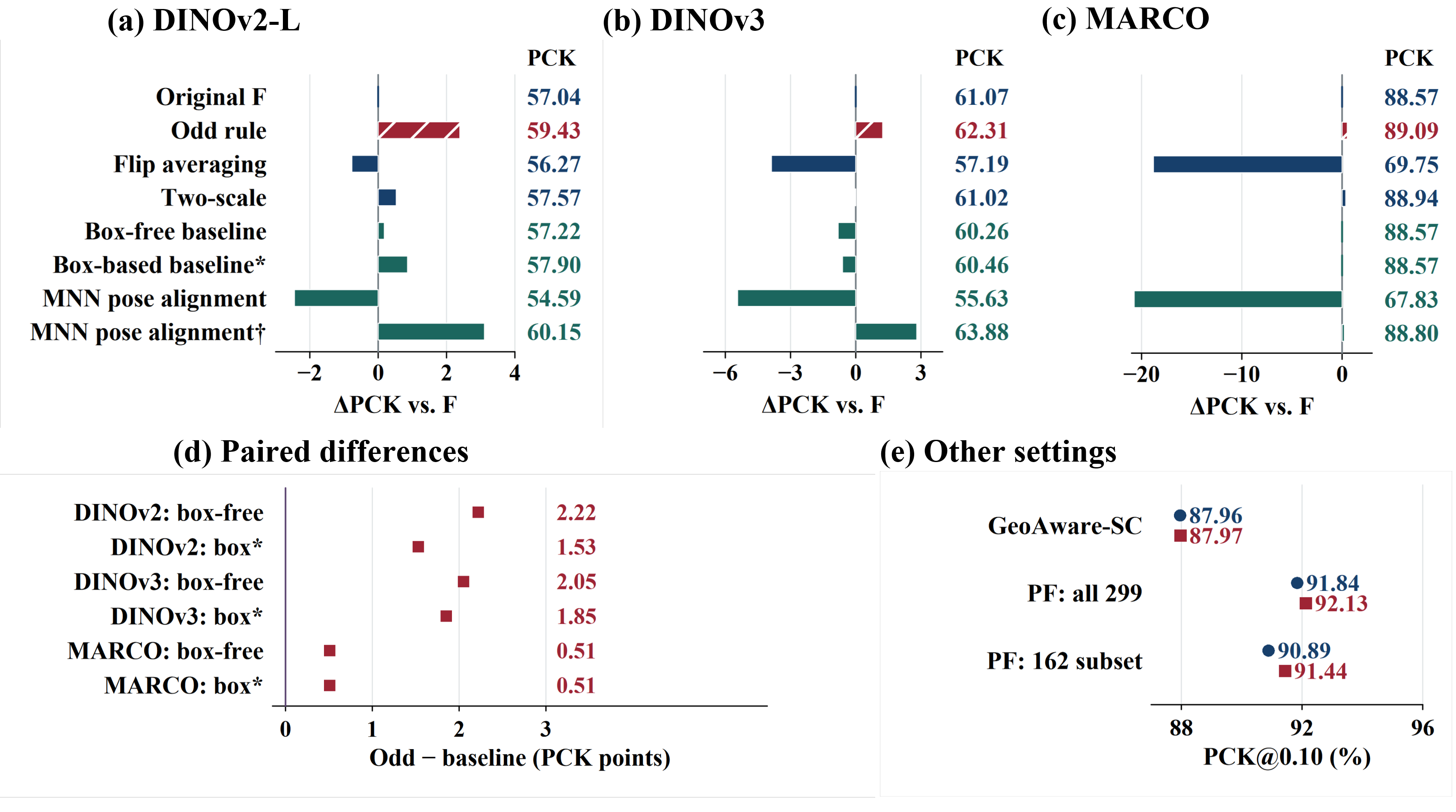}
\caption{Matching accuracy measured by PCK@0.10. In (a--c), bars show changes relative to each model's original matcher, with absolute PCK values alongside. Panel (d) shows the odd rule's PCK minus that of each position-based baseline, in pps. Panel (e) compares the original matcher and the odd rule for GeoAware-SC on SPair-71k and for MARCO on PF-PASCAL. An asterisk (*) marks baselines that use annotated boxes.}
\label{fig:7}
\end{figure}

Starting from the same candidate locations, the odd rules improve PCK over the position-based baselines, with DINOv3 using settings selected on DINOv2 validation. For DINOv2, the odd rule exceeds the validation-selected baselines by 2.22 PCK points without annotated boxes and 1.53 points with them (Fig.~\ref{fig:7}(a) and (d)). On the 4,587 box-side conflict queries, its advantage over the image-side product baseline reaches 9.09 points. We compare the two rules on all visible queries relative to the original matcher. Both rules correct the same 662 cases and introduce errors on the same 296. Beyond those shared outcomes, the odd rule corrects 3,787 cases and introduces 2,044 errors, while the image-side product baseline corrects 688 and introduces 903.

The position-based baselines selected on DINOv2 reach 57.22\% PCK for the image-side product without boxes and 57.90\% for nearest normalized box x with boxes (Table~\ref{tab:5}). Both baselines use the position score to test the even-score candidate. With those settings carried over to DINOv3, its odd rule exceeds the box-free baseline by 2.05 PCK points and its original matcher (\(F\)) by 1.24 points. All tested DINOv3 position-based baselines score at or below \(F\). For MARCO, validation selects zero weight for each added position score, so the 0.51-point gain is against \(F\). Figure~\ref{fig:7}(b)-(d) summarizes these comparisons. In Fig.~\ref{fig:7}(a)-(c), bars show changes from each original matcher and the adjacent values give absolute PCK; the axis ranges differ. An asterisk marks baselines that use annotated boxes. MNN pose alignment uses keypoint identities only for the variant marked with a dagger. Panel (d) gives PCK differences between the odd rule and each baseline.

\begin{table}[!htbp]
\centering
\small
\setlength{\parskip}{0pt}
\setlength{\tabcolsep}{4pt}
\renewcommand{\arraystretch}{1.12}
\caption{Position-Based Baselines Selected on Validation}\label{tab:5}
\begin{tabular}{@{}
  >{\raggedright\arraybackslash}p{(\linewidth - 8\tabcolsep) * \real{0.1276}}
  >{\raggedright\arraybackslash}p{(\linewidth - 8\tabcolsep) * \real{0.3734}}
  >{\centering\arraybackslash}p{(\linewidth - 8\tabcolsep) * \real{0.1581}}
  >{\centering\arraybackslash}p{(\linewidth - 8\tabcolsep) * \real{0.1984}}
  >{\centering\arraybackslash}p{(\linewidth - 8\tabcolsep) * \real{0.1425}}@{}}
\toprule\noalign{}
\begin{minipage}[b]{\linewidth}\raggedright
Model
\end{minipage} & \begin{minipage}[b]{\linewidth}\raggedright
Position-based baseline
\end{minipage} & \begin{minipage}[b]{\linewidth}\centering
Threshold / weight
\end{minipage} & \begin{minipage}[b]{\linewidth}\centering
Annotated boxes
\end{minipage} & \begin{minipage}[b]{\linewidth}\centering
\href{mailto:PCK@0.10}{\nolinkurl{PCK@0.10}}

(\%)
\end{minipage} \\
\midrule\noalign{}
DINOv2-L & Image-side product / Even-score candidate with position test & 0.1 & No & 57.22 \\
DINOv2-L & Nearest normalized box x / Even-score candidate with position test & 0.2 & Yes & 57.90 \\
DINOv3 & Transferred image-side product / Even-score candidate with position test & 0.1 & No & 60.26 \\
DINOv3 & Transferred nearest normalized box x / Even-score candidate with position test & 0.2 & Yes & 60.46 \\
MARCO & Original + position score & 0 & Depends on the position score & 88.57 \\
\bottomrule
\end{tabular}
\end{table}

How the position cue is applied matters for PCK. On DINOv2-L, using nearest normalized image x to rank the candidates yields 28.52\% PCK, while using it to decide whether to switch to the candidate proposed by the even score yields 57.05\%. Relative to the same original matches, ranking by position alone corrects 5,190 cases and introduces 30,388 errors. Testing the even-score candidate with position makes far fewer changes, with 39 corrections and 31 introduced errors. Table~\ref{tab:6-1} compares Position score only, Original + position score, and Even-score candidate with position test. The final column gives correction and introduced-error counts relative to the original matcher. For MARCO, all additive position-based baselines coincide because validation selects zero weight for each.

\begin{table}[!htbp]
\centering
\small
\setlength{\parskip}{0pt}
\setlength{\tabcolsep}{4pt}
\renewcommand{\arraystretch}{1.12}
\caption{Prespecified Position-Based Baselines}\label{tab:6-1}
\begin{tabular}{@{}
  >{\centering\arraybackslash}p{(\linewidth - 6\tabcolsep) * \real{0.3906}}
  >{\centering\arraybackslash}p{(\linewidth - 6\tabcolsep) * \real{0.1875}}
  >{\centering\arraybackslash}p{(\linewidth - 6\tabcolsep) * \real{0.2501}}
  >{\centering\arraybackslash}p{(\linewidth - 6\tabcolsep) * \real{0.1718}}@{}}
\toprule\noalign{}
\begin{minipage}[b]{\linewidth}\centering
\textbf{Baseline setting - DINOv2-L}
\end{minipage} & \begin{minipage}[b]{\linewidth}\centering
PCK (\%)
\end{minipage} & \begin{minipage}[b]{\linewidth}\centering
PCK difference: odd rule \(\mathord{-}\) baseline (pp)
\end{minipage} & \begin{minipage}[b]{\linewidth}\centering
Corrections / introduced errors (counts)
\end{minipage} \\
\midrule\noalign{}
Binary box side / Position score only & 48.27 & 11.17 & 3081/10835 \\
Binary box side / Original + position score & 57.12 & 2.31 & 1183/1116 \\
Binary box side / Even-score candidate with position test & 57.28 & 2.16 & 1850/1646 \\
Binary image side / Position score only & 47.54 & 11.89 & 1923/10314 \\
Binary image side / Original + position score & 57.04 & 2.39 & 0/0 \\
Binary image side / Even-score candidate with position test & 56.72 & 2.71 & 1018/1306 \\
Nearest normalized image x / Position score only & 28.52 & 30.92 & 5190/30388 \\
Nearest normalized image x / Original + position score & 57.04 & 2.39 & 0/0 \\
Nearest normalized image x / Even-score candidate with position test & 57.05 & 2.38 & 39/31 \\
Nearest normalized box x / Position score only & 34.56 & 24.88 & 7401/27265 \\
Nearest normalized box x / Original + position score & 57.46 & 1.97 & 3409/3041 \\
Nearest normalized box x / Even-score candidate with position test & 57.90 & 1.53 & 2379/1620 \\
Image-side product / Position score only & 29.40 & 30.03 & 6248/30664 \\
Image-side product / Original + position score & 57.00 & 2.43 & 208/248 \\
Image-side product / Even-score candidate with position test & 57.22 & 2.22 & 1350/1199 \\
\bottomrule
\end{tabular}
\end{table}

\begin{table}[!htbp]
\centering
\small
\setlength{\parskip}{0pt}
\setlength{\tabcolsep}{4pt}
\renewcommand{\arraystretch}{1.12}
\ContinuedFloat
\caption{Prespecified Position-Based Baselines (continued)}\label{tab:6-2}
\begin{tabular}{@{}
  >{\centering\arraybackslash}p{(\linewidth - 6\tabcolsep) * \real{0.4063}}
  >{\centering\arraybackslash}p{(\linewidth - 6\tabcolsep) * \real{0.1562}}
  >{\centering\arraybackslash}p{(\linewidth - 6\tabcolsep) * \real{0.2408}}
  >{\centering\arraybackslash}p{(\linewidth - 6\tabcolsep) * \real{0.1967}}@{}}
\toprule\noalign{}
\begin{minipage}[b]{\linewidth}\centering
\textbf{Baseline setting - DINOv3}
\end{minipage} & \begin{minipage}[b]{\linewidth}\centering
PCK (\%)
\end{minipage} & \begin{minipage}[b]{\linewidth}\centering
PCK difference: odd rule \(\mathord{-}\) baseline (pp)
\end{minipage} & \begin{minipage}[b]{\linewidth}\centering
Corrections / introduced errors (counts)
\end{minipage} \\
\midrule\noalign{}
Binary box side / Position score only & 53.57 & 8.74 & 2176/8798 \\
Binary box side / Original + position score & 60.32 & 1.99 & 1086/1749 \\
Binary box side / Even-score candidate with position test & 60.14 & 2.17 & 1115/1934 \\
Binary image side / Position score only & 52.99 & 9.32 & 1424/8556 \\
Binary image side / Original + position score & 61.07 & 1.24 & 0/0 \\
Binary image side / Even-score candidate with position test & 59.57 & 2.74 & 639/1965 \\
Nearest normalized image x / Position score only & 33.73 & 28.58 & 4695/28840 \\
Nearest normalized image x / Original + position score & 61.07 & 1.24 & 0/0 \\
Nearest normalized image x / Even-score candidate with position test & 61.00 & 1.31 & 7/69 \\
Nearest normalized box x / Position score only & 38.64 & 23.67 & 5715/25521 \\
Nearest normalized box x / Original + position score & 58.92 & 3.39 & 2901/4795 \\
Nearest normalized box x / Even-score candidate with position test & 60.46 & 1.85 & 1214/1750 \\
Image-side product / Position score only & 35.67 & 26.64 & 5787/28221 \\
Image-side product / Original + position score & 60.89 & 1.42 & 209/366 \\
Image-side product / Even-score candidate with position test & 60.26 & 2.05 & 697/1406 \\
\bottomrule
\end{tabular}
\end{table}

\begin{table}[!htbp]
\centering
\small
\setlength{\parskip}{0pt}
\setlength{\tabcolsep}{4pt}
\renewcommand{\arraystretch}{1.12}
\ContinuedFloat
\caption{Prespecified Position-Based Baselines (continued)}\label{tab:6-3}
\begin{tabular}{@{}
  >{\centering\arraybackslash}p{(\linewidth - 6\tabcolsep) * \real{0.3906}}
  >{\centering\arraybackslash}p{(\linewidth - 6\tabcolsep) * \real{0.1875}}
  >{\centering\arraybackslash}p{(\linewidth - 6\tabcolsep) * \real{0.2501}}
  >{\centering\arraybackslash}p{(\linewidth - 6\tabcolsep) * \real{0.1718}}@{}}
\toprule\noalign{}
\begin{minipage}[b]{\linewidth}\centering
\textbf{Baseline setting - MARCO}
\end{minipage} & \begin{minipage}[b]{\linewidth}\centering
PCK (\%)
\end{minipage} & \begin{minipage}[b]{\linewidth}\centering
PCK difference: odd rule \(\mathord{-}\) baseline (pp)
\end{minipage} & \begin{minipage}[b]{\linewidth}\centering
Corrections / introduced errors (counts)
\end{minipage} \\
\midrule\noalign{}
Binary box side / Position score only & 72.37 & 16.71 & 814/15124 \\
Binary image side / Position score only & 71.70 & 17.39 & 674/15579 \\
Nearest normalized image x / Position score only & 34.79 & 54.30 & 1958/49465 \\
Nearest normalized box x / Position score only & 41.33 & 47.76 & 2198/43926 \\
Image-side product / Position score only & 32.69 & 56.39 & 2033/51388 \\
Original + any position score

(\(\gamma = 0\)) & 88.57 & 0.51 & 0/0 \\
\bottomrule
\end{tabular}
\end{table}

For the frozen models, the comparison with pose alignment depends on whether it receives keypoint identities. The odd rule exceeds two-scale inference by 1.86 PCK points on DINOv2 in the same run and by 1.28 points on DINOv3. On DINOv3, two-scale inference reaches 61.02\% PCK and mutual nearest-neighbor pose alignment reaches 55.63\% without keypoint IDs and 63.88\% with them. With keypoint IDs, pose alignment exceeds the odd rule on both frozen models, by 1.57 points on DINOv3 (Fig.~\ref{fig:7}).

MARCO's odd rule exceeds mutual nearest-neighbor alignment with keypoint IDs, but its comparison with two-scale inference varies by query group. Over two-scale inference, the odd rule gains only 0.14 PCK points overall. On the geometry-aware subset with another visible part in the same semantic subgroup, PCK reaches 87.41\% with the odd rule and 86.84\% with two-scale inference, a difference of 0.56 points before rounding. On the 4,587 box-side conflict queries, the gap grows to 2.09 points (78.83\% versus 76.74\%) and remains 2.07 points on the 2,417 queries where all five position scores favor the counterpart. The observed PCK changes from F with two-scale inference are +0.20 and +0.17 points in those two groups, respectively. Under the original hard-peak protocol (Section 3.6), MARCO reaches 85.83\% mean per-pair PCK@0.10. Per-image PCK averages the keypoint accuracy within each image pair and then averages those pair accuracies. The published system reports 86.7\% using keypoint-derived bounding boxes and 87.2\% using Segment Anything Model (SAM) masks to restrict pseudo-label mining \cite{r3}.

Does the improvement from hard-peak reranking carry over to soft localization? Table~\ref{tab:7} compares the three center choices while keeping the final dot-product field and decoder fixed. The raw dot-product maximum (A) and the original cosine peak (B) gave the same center on 37.51\% of queries. Moving from the original cosine peak (B) to the odd-ranked peak (C) under the shared soft decoder changed 3,840 centers, corrected 1,371 predictions and introduced 909 errors at PCK@0.10. This raised per-query PCK by 0.52 percentage points (88.62\% to 89.14\%, corresponding to the 462 net corrections across 88,328 queries) and mean per-pair PCK by 0.47 percentage points (85.88\% to 86.35\%).

\begin{table}[!htbp]
\centering
\small
\setlength{\parskip}{0pt}
\setlength{\tabcolsep}{4pt}
\renewcommand{\arraystretch}{1.12}
\caption{MARCO Soft-Localization PCK@0.10 (\%) and Paired Differences (pp)}\label{tab:7}
\begin{tabular}{@{}
  >{\centering\arraybackslash}p{(\linewidth - 12\tabcolsep) * \real{0.2955}}
  >{\centering\arraybackslash}p{(\linewidth - 12\tabcolsep) * \real{0.1093}}
  >{\centering\arraybackslash}p{(\linewidth - 12\tabcolsep) * \real{0.1093}}
  >{\centering\arraybackslash}p{(\linewidth - 12\tabcolsep) * \real{0.1093}}
  >{\centering\arraybackslash}p{(\linewidth - 12\tabcolsep) * \real{0.1254}}
  >{\centering\arraybackslash}p{(\linewidth - 12\tabcolsep) * \real{0.1254}}
  >{\centering\arraybackslash}p{(\linewidth - 12\tabcolsep) * \real{0.1254}}@{}}
\toprule\noalign{}
\begin{minipage}[b]{\linewidth}\centering
\textbf{All visible queries}
\end{minipage} & \begin{minipage}[b]{\linewidth}\centering
\textbf{A}
\end{minipage} & \begin{minipage}[b]{\linewidth}\centering
\textbf{B}
\end{minipage} & \begin{minipage}[b]{\linewidth}\centering
\textbf{C}
\end{minipage} & \begin{minipage}[b]{\linewidth}\centering
\textbf{B \(\mathord{-}\) A}
\end{minipage} & \begin{minipage}[b]{\linewidth}\centering
\textbf{C \(\mathord{-}\) B}
\end{minipage} & \begin{minipage}[b]{\linewidth}\centering
\textbf{C \(\mathord{-}\) A}
\end{minipage} \\
\midrule\noalign{}
Per query & 88.41 & 88.62 & 89.14 & +0.20 & +0.52 & +0.72 \\
Per pair & 85.48 & 85.88 & 86.35 & +0.40 & +0.47 & +0.87 \\
\bottomrule
\end{tabular}
\end{table}

On the two conflict groups, odd reranking changed per-query PCK by +2.35 points for box-side conflict and +2.36 points for five-score conflict. The baseline coordinates agreed with the pinned author implementation to within 0.001 canvas pixels across 51 checked queries from 18 categories.

In supervised GeoAware-SC, stronger discrimination at the annotated parts coexists with little change in matching. In the reflected-noise setting, the odd score improves discrimination over the full descriptor by 1.653 points. Adding the weighted odd score changes the selected window center on 1,994 of 88,328 queries (2.26\%). Across the five candidates, the median within-query range (maximum minus minimum) is 0.23253 for the original similarity and 0.02311 for the weighted odd score (0.05 \(\times\) O). After soft localization, 330 predictions become correct and 315 become incorrect, shifting PCK from 87.956\% to 87.973\%. With unchanged noise, PCK remains near the baseline at 87.977\%. Averaging the original and flipped features lowers PCK by about 16 points under either setting (Table~\ref{tab:8}). The original per-image PCK of 85.592\% rounds to the authors' reported 85.6\%.

On PF-PASCAL, MARCO's odd rule has observed PCK gains of 0.29 points over the original matcher across all 299 pairs and of 0.54 points across the 162 pairs without detected overlap. These small observed gains do not establish consistent improvement on PF-PASCAL (Fig.~\ref{fig:7}(e)). In Fig.~\ref{fig:7}(e), circles denote the original matcher and squares denote the odd rule.

\begin{table}[!htbp]
\centering
\small
\setlength{\parskip}{0pt}
\setlength{\tabcolsep}{4pt}
\renewcommand{\arraystretch}{1.12}
\caption{Supervised GeoAware-SC: Noise and Score Comparisons}\label{tab:8}
\begin{tabular}{@{}
  >{\centering\arraybackslash}p{(\linewidth - 6\tabcolsep) * \real{0.3751}}
  >{\centering\arraybackslash}p{(\linewidth - 6\tabcolsep) * \real{0.1876}}
  >{\centering\arraybackslash}p{(\linewidth - 6\tabcolsep) * \real{0.1876}}
  >{\centering\arraybackslash}p{(\linewidth - 6\tabcolsep) * \real{0.2498}}@{}}
\toprule\noalign{}
\begin{minipage}[b]{\linewidth}\centering
Setting
\end{minipage} & \begin{minipage}[b]{\linewidth}\centering
PCK (\%)
\end{minipage} & \begin{minipage}[b]{\linewidth}\centering
Change from original (pp)
\end{minipage} & \begin{minipage}[b]{\linewidth}\centering
Corrections /

introduced errors (counts)
\end{minipage} \\
\midrule\noalign{}
F & 87.956 & 0.000 & 0 / 0 \\
Flip averaging, reflected noise & 71.702 & -16.254 & 2445 / 16802 \\
Odd-score rule, reflected noise & 87.973 & 0.017 & 330 / 315 \\
Flip averaging, unchanged noise & 71.845 & -16.112 & 2474 / 16705 \\
Odd-score rule, unchanged noise & 87.977 & 0.020 & 337 / 319 \\
\bottomrule
\end{tabular}
\end{table}

\subsection{Matching Errors and the Sign of the Odd Score}
\label{sec:matching-errors-and-the-sign-of-the-odd-score}

\begin{figure}[!htbp]
\centering
\includegraphics[width=\linewidth,height=0.77\textheight,keepaspectratio]{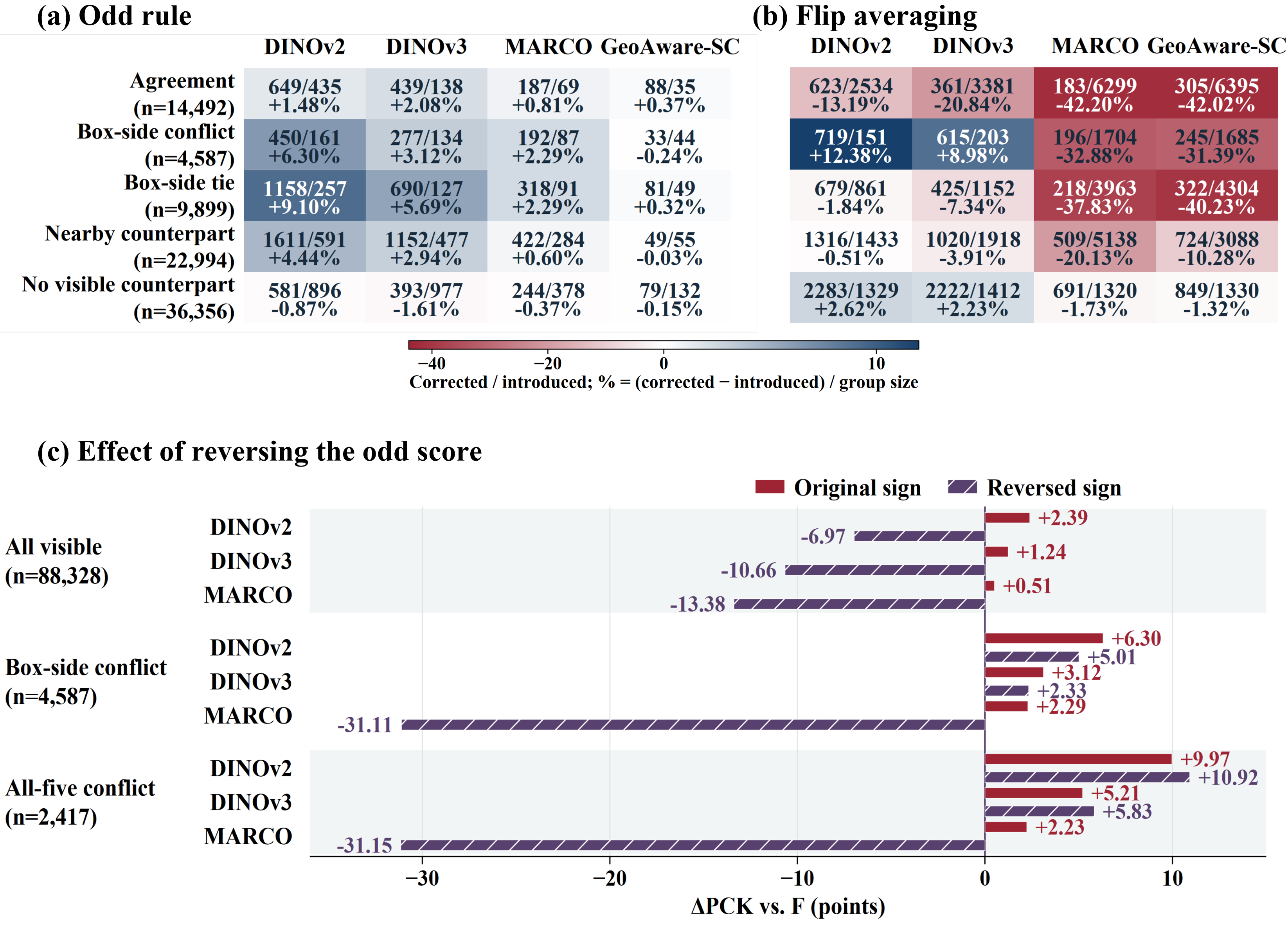}
\caption{Corrections and introduced errors by counterpart visibility and box-side preference (a--b), and matching changes after reversing the odd-score sign (c).}
\label{fig:8}
\end{figure}

The matching gains depend on whether a symmetric counterpart is visible. Figures 8(a) and 8(b) report correction and introduced-error counts relative to the original matcher from the same run, with their net difference expressed as a percentage of the group size. Across the five query groups defined in Section 3.3, the odd rule produces more corrections than introduced errors in box-side conflicts for DINOv2, DINOv3 and MARCO, while supervised GeoAware-SC loses net correct matches (Fig.~\ref{fig:8}(a)-(b)). All four lose net accuracy on the 36,356 queries without a visible annotated counterpart. Those losses offset 13.0\%, 34.7\%, 22.8\% and 77.9\%, respectively, of the net gains on queries with a visible counterpart. DINOv2 illustrates the difference in error profiles for the two mirror-based rules. Without a visible counterpart, its odd rule corrects 581 queries and introduces 896 errors, whereas flip averaging corrects 2,283 and introduces 1,329. For DINOv2, flip averaging thus gains 954 net correct matches in this group but loses 688 overall. In box-side conflicts, flip averaging exceeds the odd rule for both DINOv2 and DINOv3.

Negating the odd score on the same candidate sets leaves much of the gain in box-side conflicts for the frozen models (Fig.~\ref{fig:8}(c)). The original and negated scores yield 289 and 230 net additional correct matches for DINOv2, and 143 and 107 for DINOv3. Both signs improve on the original matcher in this group. The paired differences between the original and negated scores are 1.29 and 0.78 PCK points, respectively. These comparisons do not establish a clear advantage for the original sign in frozen-model conflicts. With the original sign, DINOv2 switches candidates on 954 conflict queries. Of these, 450 become correct, 161 become incorrect, 327 remain incorrect and 16 remain correct, leaving 466 correct predictions (48.85\%). Across all queries, DINOv3 reaches 62.31\% PCK with the original sign and 50.41\% with the reversed sign.

On five-score conflicts, the original DINOv2 and DINOv3 matchers reach 11.92\% and 16.26\% PCK, respectively (Table~\ref{tab:9}). Switching on either sign whenever the absolute odd margin exceeds 0.2 and the even tolerance is met gives 32.81\% and 27.31\% PCK, with 1,197 and 1,068 switches. The two random-control rows report expected PCK when the number of switches is matched to the positive rule. The positive rule exceeds equal-count random switching among large margins by 1.17 and 1.64 points, respectively. Neither model showed a clear advantage over this random control. On all visible queries, always accepting the even proposal when it passes the tolerance reduces PCK to 48.39\% and 45.58\%.

\begin{table}[!htbp]
\centering
\small
\setlength{\parskip}{0pt}
\setlength{\tabcolsep}{4pt}
\renewcommand{\arraystretch}{1.12}
\caption{Candidate Switches and PCK on the 2,417 Five-Score Conflicts}\label{tab:9}
\begin{tabular}{@{}
  >{\centering\arraybackslash}p{(\linewidth - 8\tabcolsep) * \real{0.4356}}
  >{\centering\arraybackslash}p{(\linewidth - 8\tabcolsep) * \real{0.1461}}
  >{\centering\arraybackslash}p{(\linewidth - 8\tabcolsep) * \real{0.1461}}
  >{\centering\arraybackslash}p{(\linewidth - 8\tabcolsep) * \real{0.1329}}
  >{\centering\arraybackslash}p{(\linewidth - 8\tabcolsep) * \real{0.1329}}@{}}
\toprule\noalign{}
\begin{minipage}[b]{\linewidth}\centering
\textbf{Switch rule}
\end{minipage} & \begin{minipage}[b]{\linewidth}\centering
DINOv2-L switches
\end{minipage} & \begin{minipage}[b]{\linewidth}\centering
DINOv2-L PCK (\%)
\end{minipage} & \begin{minipage}[b]{\linewidth}\centering
DINOv3 switches
\end{minipage} & \begin{minipage}[b]{\linewidth}\centering
DINOv3 PCK (\%)
\end{minipage} \\
\midrule\noalign{}
Always accept the even proposal & 2,336 & 41.83 & 2,400 & 33.60 \\
Random within the even tolerance & 504 & 18.37 & 345 & 18.75 \\
Random within large absolute margins & 504 & 20.71 & 345 & 19.83 \\
Positive odd margin & 504 & 21.89 & 345 & 21.47 \\
Reversed odd margin & 693 & 22.84 & 723 & 22.09 \\
\bottomrule
\end{tabular}
\end{table}

MARCO's conflict gains are more sensitive to the sign of the odd score. In box-side conflicts it gains 105 net correct predictions with the original sign and loses 1,427 when the score is negated. Where all five position scores favor the counterpart, the original sign improves PCK by 2.23 points over \(F\), while negating the score loses 753 net correct predictions (Fig.~\ref{fig:8}(c)). With the original sign, the rate of left/right swaps on the 51,972 queries with visible target counterparts falls from 23.39\% to 18.02\% for DINOv2 and from 5.03\% to 3.74\% for MARCO.

\begin{figure}[!htbp]
\centering
\includegraphics[width=\linewidth,height=0.77\textheight,keepaspectratio]{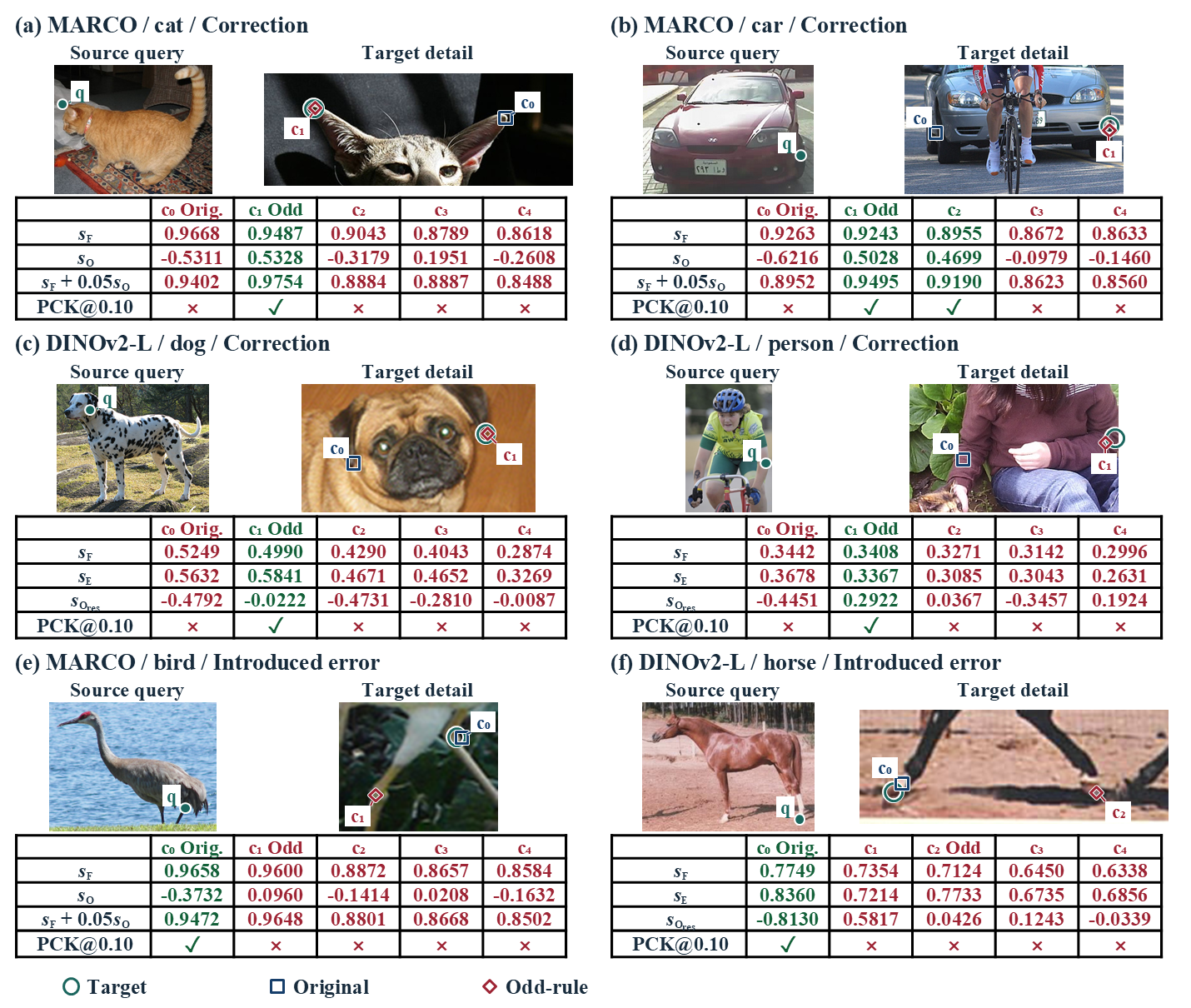}
\caption{Four corrections and two introduced errors after candidate selection. Each example compares the original and odd-rule choices among the same five peaks.}
\label{fig:9}
\end{figure}

Figure~\ref{fig:9} shows four corrections and two introduced errors among the five original candidates. The column headings identify the original and odd-rule selections, and checks and crosses show whether each candidate meets PCK@0.10. For the MARCO cat, the correct candidate initially scores 0.9487 against 0.9668 for the original peak. Adding the weighted odd score reverses their order (0.9754 versus 0.9402) and corrects the match. For the DINOv2 horse, the even proposal passes both thresholds and replaces a correct match with an incorrect one.

\subsection{Matching on AP-10K and SPair-U, and Inference Cost}
\label{sec:matching-on-ap-10k-and-spair-u-and-inference-cost}

After validation selects the parameters for each rule, the largest observed gains on the custom AP-10K pairs are 1.95 points for DINOv2 and 0.60 for MARCO. Specifically, DINOv2 improves on the original matcher most when the centered odd score tests the even-score candidate. MARCO improves most when the centered odd score is added to the original similarity (Table~\ref{tab:10}). The fixed rules also improve PCK on the added SPair-U annotations. MARCO gains 0.73 points with the additive odd score, and DINOv2 gains 0.51 with an even-score candidate and a centered odd-score test.

Supplementary Table S1 reports a separate experiment in which the matcher also receives the category and annotated part identity to construct labeled support prototypes.

\begin{table}[!htbp]
\centering
\small
\setlength{\parskip}{0pt}
\setlength{\tabcolsep}{4pt}
\renewcommand{\arraystretch}{1.12}
\caption{Original PCK (\%) and Changes (pp) on the Custom AP-10K Protocol}\label{tab:10}
\begin{tabular}{@{}
  >{\centering\arraybackslash}p{(\linewidth - 4\tabcolsep) * \real{0.3333}}
  >{\centering\arraybackslash}p{(\linewidth - 4\tabcolsep) * \real{0.3333}}
  >{\centering\arraybackslash}p{(\linewidth - 4\tabcolsep) * \real{0.3334}}@{}}
\toprule\noalign{}
\begin{minipage}[b]{\linewidth}\centering
Rule
\end{minipage} & \begin{minipage}[b]{\linewidth}\centering
MARCO
\end{minipage} & \begin{minipage}[b]{\linewidth}\centering
DINOv2-L
\end{minipage} \\
\midrule\noalign{}
Original (top-1) PCK@0.1 & 58.92 & 44.45 \\
\(s_{F} + \gamma s_{O}\) & +0.52 & 0.00 \\
\(s_{F} + \gamma s_{O_{res}}\) & +0.60 & +0.95 \\
Even-score candidate with centered odd-score test & +0.20 & +1.95 \\
\bottomrule
\end{tabular}
\end{table}

The odd rules require features from both image orientations, and sequential inference takes about twice as long as the original matcher. Relative to the original matcher, sequential odd inference raises median end-to-end latency per pair from 0.581 to 1.171 s for DINOv2 and from 0.797 to 1.639 s for MARCO. Two-scale inference takes 1.041 and 1.439 s, respectively. Batching the original and reflected inputs changes runtime little and increases peak allocated GPU memory, especially for MARCO. The timings for matching with cached features in Table~\ref{tab:11} measure matching alone.

\begin{table}[!htbp]
\centering
\small
\setlength{\parskip}{0pt}
\setlength{\tabcolsep}{4pt}
\renewcommand{\arraystretch}{1.12}
\caption{Measured Inference Cost on 108 Validation Pairs for Each Method}\label{tab:11}
\begin{tabular}{@{}
  >{\centering\arraybackslash}p{(\linewidth - 10\tabcolsep) * \real{0.1666}}
  >{\centering\arraybackslash}p{(\linewidth - 10\tabcolsep) * \real{0.2002}}
  >{\centering\arraybackslash}p{(\linewidth - 10\tabcolsep) * \real{0.1719}}
  >{\centering\arraybackslash}p{(\linewidth - 10\tabcolsep) * \real{0.1488}}
  >{\centering\arraybackslash}p{(\linewidth - 10\tabcolsep) * \real{0.1720}}
  >{\centering\arraybackslash}p{(\linewidth - 10\tabcolsep) * \real{0.1405}}@{}}
\toprule\noalign{}
\begin{minipage}[b]{\linewidth}\centering
Model
\end{minipage} & \begin{minipage}[b]{\linewidth}\centering
Method
\end{minipage} & \begin{minipage}[b]{\linewidth}\centering
Median end-to-end latency (s/pair)
\end{minipage} & \begin{minipage}[b]{\linewidth}\centering
Matching with cached features (ms)
\end{minipage} & \begin{minipage}[b]{\linewidth}\centering
Peak allocated GPU memory (GiB)
\end{minipage} & \begin{minipage}[b]{\linewidth}\centering
Estimated GFLOPs
\end{minipage} \\
\midrule\noalign{}
DINOv2-L & Original & 0.581 & 3.5 & 1.35 & 7055 \\
DINOv2-L & Odd rule, sequential & 1.171 & 13.3 & 1.36 & 14109 \\
DINOv2-L & Odd rule, batched & 1.155 & 13.2 & 1.47 & - \\
DINOv2-L & Two-scale & 1.041 & 7.0 & 1.35 & 12378 \\
MARCO & Original & 0.797 & 38.5 & 1.84 & 7359 \\
MARCO & Odd rule, sequential & 1.639 & 117.6 & 1.92 & 14719 \\
MARCO & Odd rule, batched & 1.642 & 118.6 & 2.46 & - \\
MARCO & Two-scale & 1.439 & 89.7 & 1.84 & 12929 \\
\bottomrule
\end{tabular}
\end{table}

Requesting a reflected pass only for ambiguous queries reduces the saving once several queries share an image pair. In validation replay, the lowest-quarter gate selected 25.5\% of DINOv2 queries and 25.2\% of MARCO queries, but 72.1\% and 74.2\% of the 5,384 image pairs required the extra pass. DINOv2 validation PCK was 55.38\% using the original matcher, 56.39\% using the gate and 57.40\% using the odd rule throughout. MARCO reached 90.16\%, 90.74\% and 90.71\%, respectively. Weighting the measured original and sequential-odd median times by these pair frequencies gives estimated savings of 14.1\% for DINOv2 and 13.3\% for MARCO. These estimates exclude gating overhead and cross-pair image reuse.

\section{Discussion}
\label{sec:discussion}

On SPair-71k, frozen DINOv2's raw odd descriptor reached 70.66\% left/right discrimination accuracy across the 28,978 eligible queries, close to the 70.26\% obtained by matching normalized horizontal box coordinates (Table~\ref{tab:1}). The two scores behaved differently when the box-side cue gave both parts equal scores. The descriptor reached 70.25\% accuracy on those queries, compared with 60.06\% for the coordinate baseline. Yet when all five position scores favored the opposite part, the descriptor's accuracy fell to 2.98\%. The drop from box-side agreement to conflict also appeared within each viewpoint group when we compared the same part pair in the same target image. Reporting agreement and conflict separately reveals when the overall score gives a misleading impression of left/right discrimination.

The readout also matters. Within MARCO, the backbone and adapters supply the same input to bilinear interpolation and learned upsampling. The two outputs have the same grid size, yet conflict accuracy differs sharply for both full and odd descriptors. A comparison of representation quality should therefore identify the feature map from which similarities were sampled.

The full test set and the five-score conflict subset give different answers about the direction of the odd score. Across all visible queries, DINOv2 reaches 59.43\% PCK with the original sign and 50.08\% with the reversed sign. Among the 2,417 queries where all five position scores favor the counterpart, the original sign yields 21.89\% PCK with 504 switches and the reversed sign yields 22.84\% with 693 switches (Table~\ref{tab:9}). With the count fixed at 504, random switching among queries that pass the even tolerance and the absolute odd-margin threshold gives an expected 20.71\% PCK. The advantage over random switching is unclear on this subset. A gain over the original peak alone therefore gives little evidence that the score identified the correct side. Identifying this subset requires the target annotations. In supervised GeoAware-SC, the weighted odd score spans a much smaller range than the original similarity scores, so few window centers move and corrections nearly balance introduced errors. The number of image pairs requiring a reflected view also determines the cost of applying the rule.

Matching the final grid size tests which distinctions remain readable after pooling. Pooling also changes filtering and spatial detail, and the shared extraction holds training history fixed. The matched decoder comparison addresses a different decision. Its C versus B contrast measures the consequence of the odd vote among the available cosine peaks. The B versus A contrast also changes descriptor normalization, source point sampling and the set of eligible peak locations.

\section{Limitations}
\label{sec:limitations}

The five position scores are related and incomplete. Matching the target and part pair leaves source pose and appearance variable. Groups defined by the sum of signs differ in composition and do not form independent measurements of continuous severity. Negating the score is not an exchangeable random control. The five-candidate rule has limited coverage, and centered \(O_{res}\) is odd only if \(\mu(Mp) = - \mu(p)\).

Supervision, adapters, training history and extraction geometry differ across models. Translations within each fixed reflection convention preserve relative object side. Comparisons across the primary, translation and architectural tests also differ in extraction geometry and can use different reflection conventions. Perturbations with the same Frobenius norm have different effects on downstream performance, including large losses with the original descriptor \(F\). These interventions therefore do not separate effects on part discrimination from effects on overall matching, and they do not test rotary positional embeddings. The pose-alignment baseline can receive keypoint identities, and supervised GeoAware-SC uses window soft argmax. The three-arm comparison evaluates the released Gaussian-weighted decoder on this study's features, images and preprocessing. The equivalence check applies to the sampled author-operator outputs. A separately trained or differently masked published system can have a different PCK. Odd inference is slower than two-scale inference. The feature upsampling and candidate selection experiments measure discrimination and matching gains in separate comparisons.

SPair outputs informed development and follow-up analysis; SPair-U shares its images. PF auditing cannot exclude all near duplicates or pretraining overlap. We apply the matching settings selected on DINOv2 to DINOv3. Custom AP-10K pairs differ from public intra-species, cross-species and cross-family evaluation~\cite{r4}. Queries assigned to different targets can still share a source image.

The FLOP estimates omit normalization, interpolation, sampling and elementwise work. The timing does not measure amortized throughput. We lack the alternative-window predictions needed to reconstruct GeoAware-SC's full reversed-score PCK.

\section{Conclusion}
\label{sec:conclusion}

Overall discrimination accuracy does not show whether a descriptor remains reliable when position favors the wrong bilateral part. We addressed this problem by comparing position agreement and conflict within the same target images and annotated part pairs. Decomposing the response to horizontal reflection allowed us to evaluate full, even and odd descriptors under the same conditions. In the matched comparison, odd accuracy fell from 86.66\% to 19.63\% for frozen DINOv2 and from 95.27\% to 86.16\% for MARCO. With the backbone and adapters fixed, learned upsampling improved full and odd discrimination in the translation comparison. The pooled learned features distinguish the parts more accurately than the native features in both conflict groups, even at the same 60 \(\times\) 60 output size. We then tested whether odd scores improve coordinate selection. The rules improved on the original matchers overall for DINOv2 and DINOv3 and on conflict queries for MARCO. With the soft decoder fixed, MARCO reached 85.88\% mean per-pair PCK@0.10 when the window was centered on the highest cosine-similarity peak. Reranking the same five peaks with the odd score raised this to 86.35\%, a gain of 0.47 pps. In supervised GeoAware-SC, using the odd score to select the localization window center corrected 330 matches while introducing 315 errors. These results show why evaluations should distinguish part discrimination from matching performance: accuracy under position conflict tests whether a descriptor favors the correct part when a position cue does not, while correction and introduced-error counts reveal the consequences of using that score to choose a match.

\section*{Data and Code Availability}
\label{sec:data-and-code-availability}

The images and pretrained weights are available from their original distributors.

\section*{Declaration of generative AI and AI-assisted technologies in the writing process}
\label{sec:declaration-of-generative-ai-and-ai-assisted-technologies-in-the-writing-process}

The authors used ChatGPT for language editing throughout the manuscript. The author takes full responsibility for the content of this article.

\FloatBarrier

\end{document}